\documentclass[11pt, a4paper, oneside]{article}
\pdfoutput=1

\usepackage{braket}
\usepackage{natbib}
\usepackage[nottoc]{tocbibind}
\usepackage{hyperref} 
\usepackage{amsmath,amssymb}

\usepackage{times}

\usepackage{soul,float} 

\newcommand{\argmax}{\mathop{\mathrm{argmax}}}

\newcounter{authorc}

  \usepackage[utf8x]{inputenc}
  \usepackage[ngerman,english]{babel}
  \usepackage{microtype}
  \usepackage{fancyvrb}
  \usepackage{pdfpages}
  \addto\extrasenglish{%

  }

  \usepackage{graphicx}
  \usepackage{dashrule}
  \usepackage{geometry}
  
\usepackage{lscape,booktabs,longtable,tabularx}
    \usepackage{pdflscape}
    \usepackage{afterpage}
  \makeatletter
    \newcommand{\chapterauthor}[1]{%
    {\parindent0pt\vspace*{-25pt}
    \linespread{1.1}\large\scshape#1
    \par\nobreak\vspace*{35pt}}
    \@afterheading
  }
  \makeatother
  \usepackage{amsmath}

  \usepackage[small,sc,labelsep=quad]{caption} 
  \usepackage{bookmark}

  \usepackage{enumitem}
   \newcommand{\TitleFrame}[2]{%
   \fboxrule=\FrameRule \fboxsep=\FrameSep
   \vbox{\nobreak \vskip -0.7\FrameSep
     \rlap{\strut#1}\nobreak\nointerlineskip
     \vskip 0.7\FrameSep
     \noindent\fbox{#2}}}
   {\endMakeFramed}

\usepackage{setspace}
\usepackage{import}
\newcommand{\longTitle}{A time resolved clustering method revealing longterm structures and their short-term internal dynamics}

\newcommand{\theAuthors}{
Jonas I. Liechti
\\ Institute for Integrative Biology, \\
ETH Z\"urich, 8092 Z\"urich, \\
Switzerland
\and
Sebastian Bonhoeffer
\\ Institute for Integrative Biology, \\
ETH Z\"urich, 8092 Z\"urich, \\
Switzerland
}

\begin{document}
  
  \title{\longTitle}
  \author{\theAuthors}
  \date{\today} 
  \maketitle
  \begin{abstract}
	The last decades have not only been characterized by an explosive growth of data, but also an increasing appreciation of data as a valuable resource.
	Their value comes with the ability to extract meaningful patterns that are of economic, societal or scientific relevance.
	A particular challenge is the identification of patterns across time, including those that might only become apparent when the temporal dimension is taken into account.
	Here, we present a novel method that aims to achieve this by detecting dynamic clusters, i.e.~structural elements that can be present over prolonged durations.
	It is based on an adaptive identification of majority overlaps between groups at different time points and accommodates the transient decompositions in otherwise persistent dynamic clusters.
	Our method enables the detection of persistent structural elements with internal dynamics and can be applied to any classifiable data, ranging from social contact networks to arbitrary sets of time stamped feature vectors. 
	It represents a unique tool to study systems with non-trivial temporal dynamics and has a broad applicability to scientific, societal and economic data.


\end{abstract}

  \clearpage
  \section{Introduction}
With digitalization penetrating all aspects of life we are witnessing an explosive growth of data.
Data clustering~\citep{kaufman2009finding}, i.e.\  a categorization of data sources into different groups, is one of the most popular approaches to harvest knowledge from this deluge of data.
In countless applications data clustering has shown to reveal latent yet meaningful structures.  
Clustering can be applied to both non-relational data (information about individual data sources) and relational data (information about the relation between data sources).
In non-relational data, clustering aims to group data sources based on some measure of similarity. 
In relational data, clustering - also called community detection - focuses on identifying sets of data sources that are more densely connected within, as compared to between, sets.

The bulk of methods for cluster detection both in non-relational~\citep{jain1999data} and in relational data~\citep{fortunato2010community, fortunato2016community} has been developed for static datasets.
However, one particular aspect of the ever growing amount of available data is the temporal dimension. 
In temporal data, each data source might contribute several data points to the dataset, each with a different time stamp.
Including this temporal information allows to delineate the evolution of a system.
The temporal information can be crucial for the understanding of observed patterns, as many systems are intrinsically dynamic;
any observed state can only be explained in light of the history of the system.
Some of the pertinent examples highlighting the importance of temporal dynamics are social media~\citep{chakrabarti2006evolutionary}, mobile subscriber networks~\citep{palla2007quantifying} or co-authorship relations~\citep{backstrom2006group,rosvall2010mapping}.

In the last decade and a half, considerable efforts were made to extend static methods to time-stamped data and develop new ones capable to cope with temporal data.
Such methods are commonly referred to as \textit{evolutionary clustering}, a term shaped by~\citet{chakrabarti2006evolutionary}, or \textit{dynamic community} detection in the context of social network analysis.
A topical overview can be found in the review by~\citet{dakiche2019tracking}.

A common representation of time stamped data is in the form of a sequence of snapshots, with each snapshot being an aggregation of data points over a certain amount of time, e.g. \textit{per day}, \textit{per month} or \textit{per year}.
In a single snapshot each data source is maximally present with a single data point.
This data point is the result of an aggregation, if a data source contributes several data points to a single snapshot.
For relational data such representation is also called \textit{time-window graphs}~\citep{holme2015modern}.
The advantage of this approach lies in the representation of temporal data in the form of a series of static datasets that can be analyzed using the rich tool-set from traditional clustering analysis.
The drawback is the loss of all temporal information about the data sources within the aggregation windows.
Several approaches to include temporal information adapt either the snapshot representation, like the creation of joint graphs from two snapshots \citep{palla2007quantifying}, or the measures from static clustering~\citep{dinh2009towards,sun2007graphscope}, or both~\citep{mucha2010community}.
Another option is to define rules to combine a sequence of clusterings resulting form static methods applied to each snapshot.
We will refer to this approach as \textit{ad hoc evolutionary clustering}.
It can be considered to be a more general approach due to its independence on the clustering method used.
Ad hoc evolutionary clustering methods are by definition applicable both to non-relational and to relational data.

Evolutionary clustering methods define \textit{dynamic clusters} (DCs), i.e.\ clusters that might persist over several snapshots, based on rules that relate clusters between time points.
Careful thought and consideration should be given to the definition of those rules and their underlying principles.
Ideally, the concept of a DC is defined \textit{a priori}, such that the set of rules is an implementation of the concept and not the other way around.

Here, we propose an ad hoc evolutionary clustering method to detect DCs in temporal data.
Our method is highly flexible, as the only requirement for its application is a time-series of clusterings, which can be generated by any clustering method applied to relational or non-relational data, including overlapping community detection methods for relational data, such as the one by~\citet{palla2005uncovering}.
Our framework utilises majority as basis to detect DCs.
It features a rule-set that allows to adapt the temporal scale at which processes are deemed relevant for the dynamic cluster structure.
As a result, the framework allows to capture and study short-lived changes, e.g.~natural fluctuations, small perturbation or small-scaled processes within the clusters, such as fission-fusion dynamics~\citep{aureli2008fission}, without loosing track of the dynamics at longer time scales. 
Finally, it is applicable to ``live'' datasets, i.e.~with continuously generated data.

In the following, we dissect the life-cycle events of a DC in the context of a sequence of snapshots.
We specify a set of properties that explicitly define what we consider to be a robust DC.
Then, we present a set of rules, along with an algorithmic procedure, to detect DCs. 
In addition, the functioning of the novel framework is illustrated by means of synthetic examples and two exemplary use cases: the voting behaviour of US senators throughout the history of the US congress and the social structure over several generations of a population of free-ranging house mice.

\section{The life-cycle of a dynamic cluster}
\label{method:sec:lifecycle}

\begin{figure}[h!]
\centering
\makebox[\textwidth][c]{\includegraphics[width=0.8\textwidth]{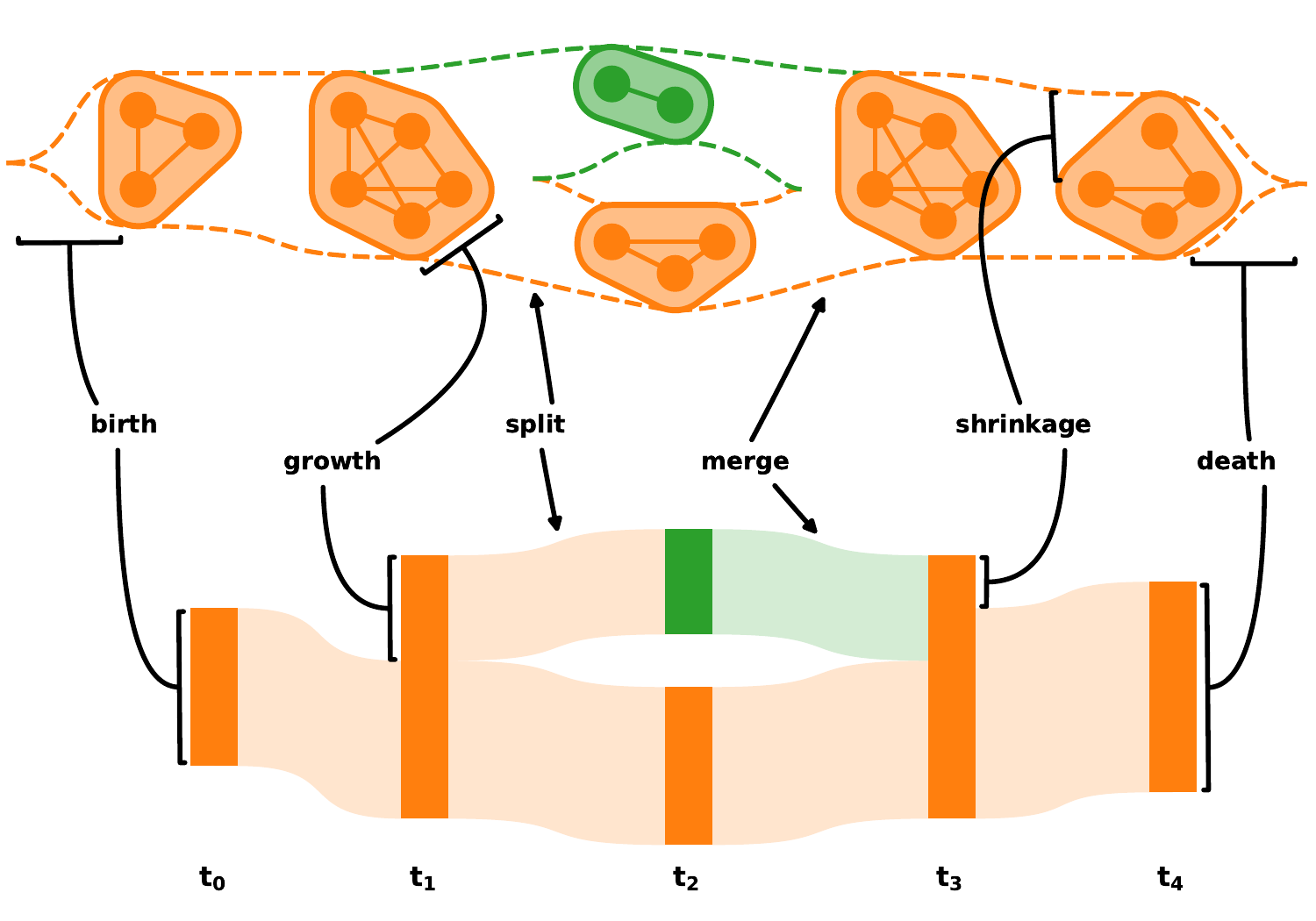}}
\caption{\small%
	\textbf{Illustration of events in the life-cycle of a dynamic cluster (DC).}
	\textit{Upper part}: DC with individual members in a sequence of snapshots of a time stamped dataset.
	In non-relational data, clusters originate from similarity measures between data points, as highlighted by the position of each data point within a snapshot.
	In relational data, the position is irrelevant and clusters correspond to (more densely) connected data sources (nodes).
	Relations between data sources are illustrated by lines between them.
	Enclosed sets of nodes correspond to clusters, with the colour implicating the associated DC.
	The dashed lines are visual guidelines to track the present DCs through time.
	\textit{Lower part}: Cluster based representation of a DC using an Alluvial diagram. 
	Each block corresponds to a cluster. 
	The height of a block represents the cluster size, the width has no particular meaning.
	The flows between blocks illustrate how the data sources redistribute between time points.
	The difference between block height and summed height of in- and out-flows corresponds to the number of introduced and removed data sources, respectively.
  \label{method:fig:lifecycle}
}
\end{figure}

In a sequence of snapshot representations of temporal data, DCs consist of a time-series of sets of data sources.
We will refer to these data sources as the \textit{members} of the DC.
Changes in this time-series of members will determine the life-cycle of the DC.
Such changes can be classified into six elementary events: \textit{birth}, \textit{death}, \textit{growth}, \textit{shrinkage}, \textit{split} and \textit{merge}, illustrated in~\autoref{method:fig:lifecycle}.
The life-cycle of a DC can be described in general terms and even before exact rules are defined how a DC is identified (see~\autoref{lifecycleSI}). 
However, robustly linking observed patterns to these life-cycle events requires a set of explicitly defined rules.
In combination with an algorithmic application procedure they define an ad hoc evolutionary clustering method. 
Before we proceed to the presentation of our novel method, we present the properties of a dynamic cluster that we define as robust DC and that will figure as underlying principles to the set of rules for our method.

\section{Definition of a robust dynamic cluster}

A simple and intuitive principle to link clusters over time, in order to build a DC, is a majority based association, where the biggest fraction of the members of a cluster is followed to some other point in the past or future.
We use bijective majority based relations, i.e.\ clusters from neighbouring time points reciprocally hold each others biggest fraction of members, as a first criterion to identify DCs.

\begin{figure}[h!]
\centering
\makebox[\textwidth][c]{\includegraphics[width=1.1\textwidth]{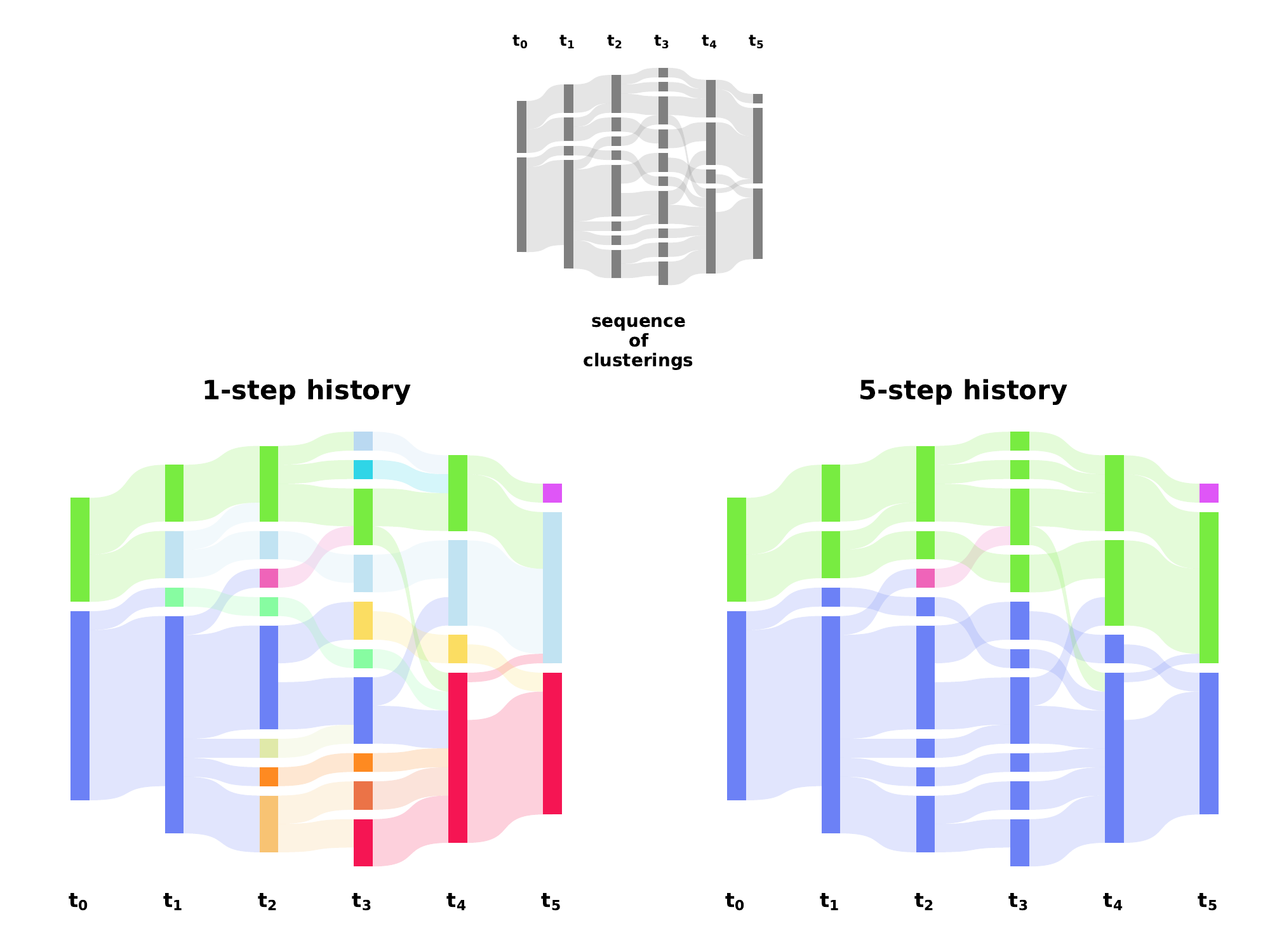}}
\caption{\small%
	\textbf{Illustration of how continuity in expression of a persistent dynamic cluster (DC) affects the observed structural dynamics.}
	\textit{Center top}: Alluvial diagram showing the time-series of clusters in a sequence of snapshots from a synthetic dataset.
	\textit{Left}: The same alluvial diagram but with each colour representing a DC.
	The DCs are formed based on the condition of continuous presence throughout their life-span.
	The association criterion for clusters is based on a simple bijective majority rule between each neighboring time point, i.e.~two clusters from neighbouring time points are associated to the same DC if they contain the majority of each other's members.
	\textit{Right}: The same alluvial diagram, but with each colour representing a DC.
	In this case, the condition on continuity is relaxed.
	A particular DC must be detectable within a limited number of time points - here set to five.
	The association criterion is still based on a bijective majority rule, but generalized to distant time points.
	For a detailed description of the method see~\autoref{method:sec:method} or refer to~\autoref{methodSI:tbl:algoDesc} from the supplementary material. 
  \label{method:fig:mem_nomem}
}
\end{figure}

An additional challenge, despite the establishment of linking relations between neighbouring snapshots, are the concepts of persistence and continuity.
Intuitively, we tend to identify DCs as sets of data sources that appear as related clusters over several consecutive time-points.
If they only appear together in a cluster every other time, identifying them as members of a DC becomes more dubious.
While in both cases the DC shows persistence over time, the latter lacks continuity. 
Many real systems are prone to produce discontinuous but persistent structures.
We will follow the nomenclature of~\citet{rosenberg2007v} and use the term \textit{homogeneous} discontinuities to describe cases where a DC transiently decomposes into sub-units.
A variety of dynamic patterns fall into this classification, most notably the fission-fusion patterns well-studied in social systems~\citep{aureli2008fission}.
We identify two elementary patterns in DCs with homogeneous discontinuities:~\textit{splintering} and \textit{transitioning}.
As splintering, we denote events where members of a DC are temporally split into several sub-clusters.
A transitioning event is defined by a series of points, during which the members gradually attach to a splinter sub-cluster, until all members have transitioned and the growing splinter cluster effectively becomes the initial DC.
Both events must be transient, as otherwise the initial DC can not be considered to persist.
The definition of an adequate time-scale limit at which splintering and transitioning events can be considered transient, must be defined on a case-by-case basis.
\autoref{method:fig:mem_nomem} illustrates how two extreme choices in the time-scale limit affect the detection of DCs.
We add to the list of desirable features the capability to track clusters over time in spite of homogeneous discontinuities.

\begin{figure}[h!]
\centering
\makebox[\textwidth][c]{\includegraphics[width=0.8\textwidth]{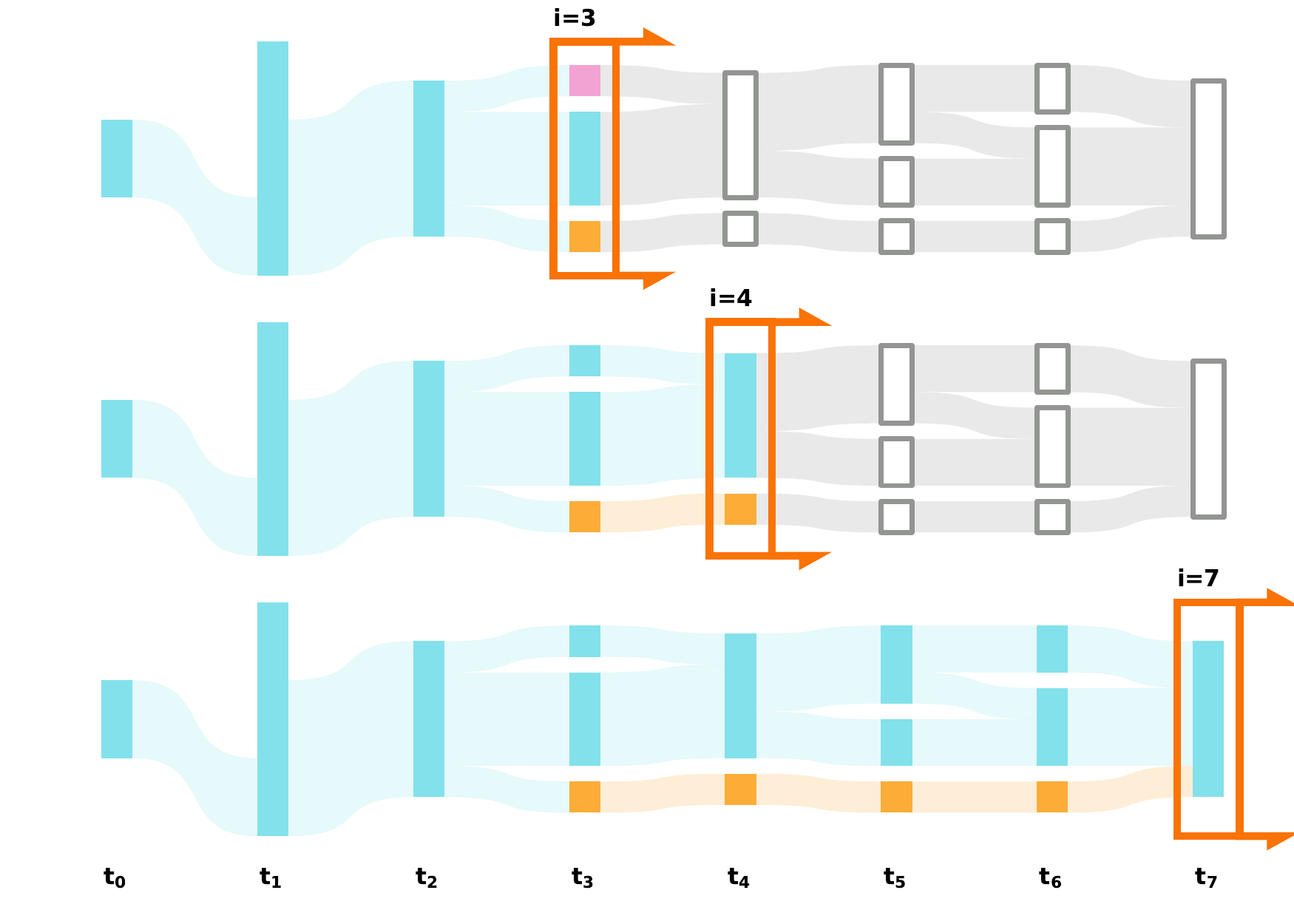}}
\caption{\small%
	\textbf{Illustration of different stages of the progressive detection algorithm.}
	The status of the progressive algorithm at the time of observation is indicated using an orange arrowed box and specified by the time index $i$.
	The defined features are illustrated as follows:
	\textit{Majority based}: Associations of clusters at $t_i$ to existing DCs are based on a bijective majority match with clusters from earlier time points (see~\autoref{method:sec:method} for details).
	\textit{Progressiveness}: Gray elements correspond to future data and are, therefore, irrelevant to the DC detection algorithm.
	\textit{Robustness against turnover}: Strong size fluctuations of a DC, (e.g.\ the growth from $t_1$ to $t_2$ by over $100\%$ and shrinkage between $t_2$ and $t_3$) should not affect the identification of a DC.
	\textit{Structural consistency}: The algorithm should generate a DC structure that coincides with the present clustering.
	Each cluster at the current stage of the algorithm has an individual colour, as illustrated by step $i=3$.
	\textit{Splinter-Resilience}: From the three DCs identified at $i=3$, the top most cluster is retrospectively identified at $i=4$ as a splinter sub-cluster, and thus, is re-integrated into the blue DC. 
	\textit{Transition-Resilience}: At stage $i=5$ the blue DC splinters.
	Until stage $i=7$ the splinter completely absorbed the rest of the DC, effectively recombining it.
	\textit{Transient discontinuity}: Among the two smaller DCs at $i=3$, only the upper (pink) classifies as a splinter.
	The orange one remains separated considerably longer, until $t_7$, from the blue DC and is therefore counted as a DC on its own.
  \label{method:fig:requirements}
}
\end{figure}

With these clarifications at hand, we postulate that the following set of features should define a DC and figure as a basis for the implementation of a detection algorithm (see \autoref{method:fig:requirements} for a visual representation):
\begin{description}
	\item[Majority based]
		DCs must be identified using a bijective majority based association criterion between clusterings from different points in the time-series.
		Clusters that reciprocally hold each others biggest fractions of members, should belong to the same DC.
	\item[Progressive]
		A dynamic clustering must be based on existing data and be capable to incorporate newly generated data into the existing structure, i.e.\ DCs must be progressively detectable on a live dataset.
		This feature is equivalent to the condition that the dynamic clusters, at any point in time, can only depend on data from that and, potentially, any previous time point.
	\item[Robust against high turnovers]
		The dynamic structure should show minimal dependency on the introduction and disappearance of data sources in the dataset.
		It ensures that the DC structure only follows structural changes and is not dominated by the turnover of data sources in the dataset.
	\item[Structurally consistent]
		The most recent DC structure should always coincide with the clustering detected in this snapshot.
		Structural consistency is of particular importance in a live dataset, because the last (or current) snapshot continuously changes with incoming data.
		However, this requirement does not exclude the possibility that the clustering deviates from that of DCs, at points previous to the last point of observation.
	\item[Resilient against homogeneous discontinuities]
		Persistent structures showing homogeneous discontinuities should still be identified as a DC.
		Homogeneous discontinuities are transient decompositions of two different types:
		\begin{description}
			\item[\textit{Splintering:}] Decomposition into sub-clusters.
			\item[\textit{Transitioning:}] Emergence of a sub-clusters that gradually absorbs the majority of members from the DC.
		\end{description}
	\item[Sensitivity to time-scale of discontinuity]
		The time-scale at which discontinuities are considered to be transient is context specific, and thus, needs to be adaptable.
\end{description}

\section{A novel ad hoc evolutionary clustering method}
\label{method:sec:method}
For an in-depth description and definition of each term introduced in this section refer to~\autoref{detailsSI}.

\subsection{Relating neighbouring snapshots}
The implementation of a majority based identification of clusters from different time points will determine the mechanistic definition of a DC.
Practically, we first assess the similarity of two clusters from consecutive snapshots.
To do so, we use the fraction of shared members, only considering resident members, i.e.\ data sources that contribute a data point to both snapshots.
We divide the size of the intersection of two clusters by the number of resident individual in one and the other cluster, respectively.
This yields two similarity measures that are unaffected by member turnover.
These measures represent the fraction of members from one cluster present in the other, and vice-versa.
They can be considered as non-symmetric variations of the well known Jaccard index~\citep{jaccard1901distribution}.
Based on these similarity measures, we identify the clusters in the neighbouring snapshots that are most similar. 
Since our similarity measure is not symmetric, we use the term \textit{mapping relation} whenever we follow the majority forward in time, and \textit{tracing relation} for a time backward direction.
We will use a \textit{bijective majority match}, i.e.~a cluster is the tracing cluster of its own mapping cluster, as a condition to associate two cluster from consecutive snapshots to each other and, ultimately, to the same DC.
Note that a posterior cluster could trace back to more than one prior cluster, and the same holds true in the other direction.
Therefore, we consider sets of clusters when identifying bijective majority matches.

\subsection{Generalisation to relations between distant time points}

What renders this approach non-trivial, and will allow us to implement the remaining required features, is a generalisation of mapping and tracing relations to a measure between snapshots from distant time points.
Concretely, for each cluster we will identify the earliest set of clusters with which it forms a bijective majority match and try to associate the cluster to the same DC as the clusters in this set.
To generalise, we apply the matching between consecutive time points iteratively.
Following tracing relations back over several snapshots we can construct what we call a \textit{tracing path}, and a \textit{mapping path} for the inverted direction.
A cluster forms a bijective majority match with a set of clusters from an earlier time point if, at some depth or recursion, the tracing path of the cluster equates to this set and the mapping path from the set equates a set with the later cluster as its unique member.
There is, in principle, no restriction on the number of time points over which a bijective majority match occurs.
We include the possibility of such a restriction in the form of a parameter, determining the maximal recursion depth of the tracing and mapping paths.
This parameter allows, on the one hand, to set a limit to the duration of within DC processes and, on the other hand, to study the types of transient decompositions present in the data by comparing DC structures generated under different restrictions for this maximal length.

We consider a cluster to hold its own majority, hence a cluster always forms a bijective majority match at least with itself.
If a cluster forms a bijective majority match with sets of clusters from earlier snapshots, we associate the cluster to the DC of the earliest \textit{source set}, i.e.~the earliest set that forms a bijective majority match with the target cluster and contains exclusively clusters associated to the same DC.
The combination of all clusters in the tracing and mapping paths connecting the target cluster to this source set can be seen as the sequence of sets of clusters along which the identity of a DC can propagate through time, and will be referred to as the \textit{identity flow} of the target cluster.

An identity flow spanning over more than two snapshots can enclose sequences of cluster sets (see blue clusters in~\autoref{method:fig:algoCompact}).
These embedded sequences are necessarily shorter than the maximal length of an identify flow and satisfy our conditions for transient homogeneous discontinuities.
We thus identify clusters belonging to embedded sequences as marginal parts of the embedding DC.


\subsection{Definition of a dynamic cluster}

We define a DC as a sequence of sets of clusters, where each cluster either, (i) has an identity flow with a source set contained in the DC, (ii) is contained in an identity flow of a cluster belonging to the DC, or (iii) is part of the clusters marginalised by the identify flow of other clusters belonging to the DC (for a mathematical definition refer to~\autoref{detailsSI}).

The identification of the source set of a cluster is conditioned on the maximal length of a bijective majority match which must be determined \textit{a priori} through a parametrisation of the method.
According to this definition, a minimal configuration of a DC might consist of a set containing a single cluster only.

\subsection{Algorithmic procedure}
\label{method:sec:dcdalgh}
\begin{figure}[h!]
\centering
\makebox[\textwidth][c]{\includegraphics[width=0.7\textwidth]{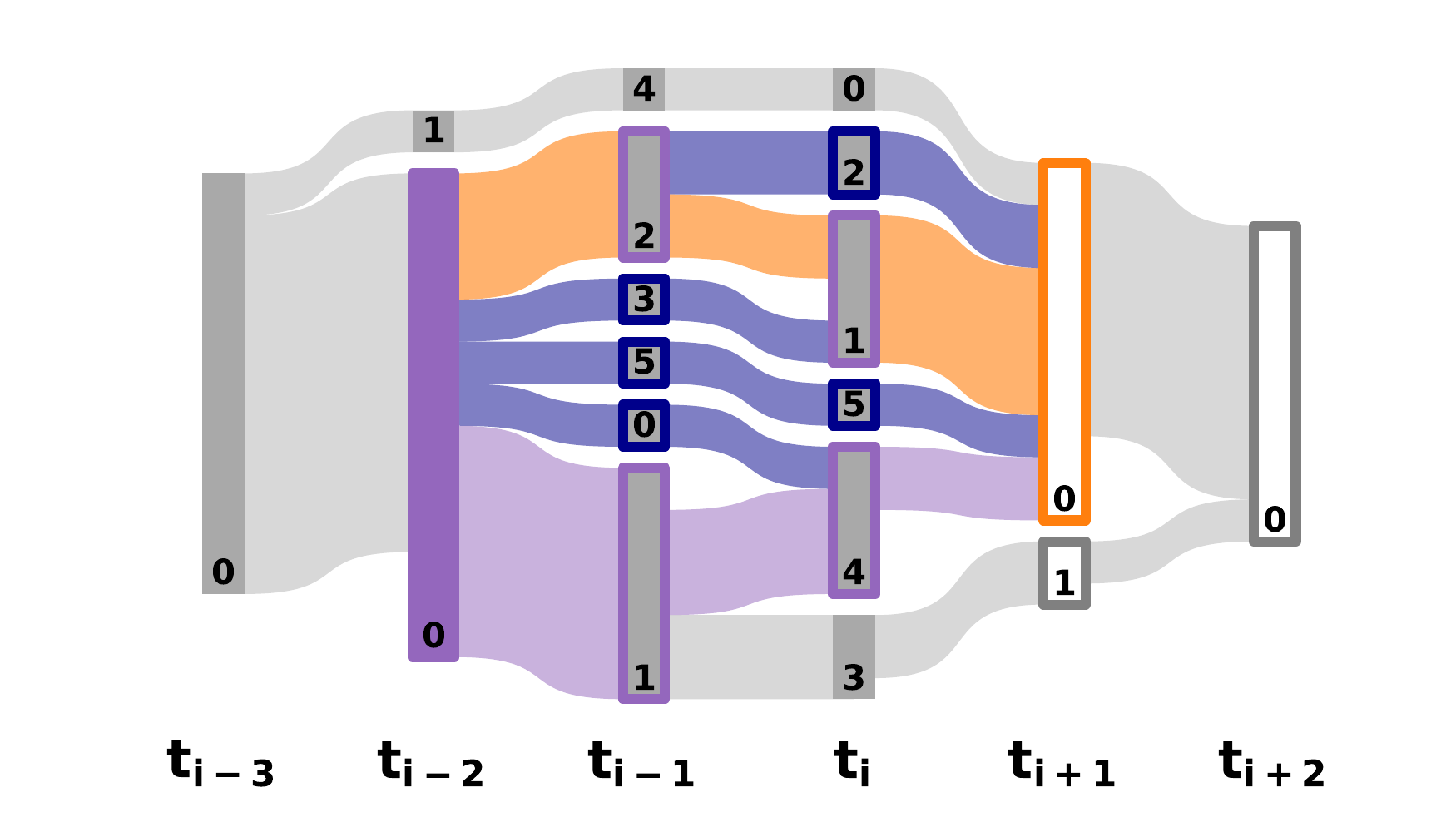}}
\caption{\small%
	\textbf{Illustration of a DC-cluster association with a 3-step history.}
	Framed in orange is the target cluster at $t_{i+1}$ that will be associated to the DC of the source set of clusters at $t_{i-1}$, coloured in purple.
	The source set is defined as the earliest candidate set for a bijective match, given the history parameter and consists of a single cluster in the current example.
	The association is based on a bijective majority match between the target cluster and the source set.
	The fluxes, highlighted in orange, show the tracing path of the target cluster, while those highlighted in purple describe the mapping path from the source set.
	Together, the purple and yellow fluxes are called the identity flow (see~\autoref{methodSI:tbl:algoDesc} for further details).
	Fluxes coloured in blue designate tracing or mapping paths that are attached exclusively to clusters from the identity flow.
	By construction, these fluxes do not contribute to the identity propagation of a DC and are thus called marginal flows.
	The ensemble of involved clusters, indicated by coloured frames, will be associated to the purple DC as a result from the illustrated bijective match.
  \label{method:fig:algoCompact}
}
\end{figure}

Refer to~\autoref{method:fig:algoCompact} for a visual illustration of the algorithmic procedure, or to~\autoref{methodSI:tbl:algoDesc} of the supplementary material for an in depth step-by-step description.

The algorithmic procedure defining our ad hoc evolutionary clustering method consists in passing through the sequence of snapshots, starting at the earliest time point and performing two distinct tasks for each cluster in the current snapshot:
\begin{enumerate}
	\item Associate the cluster and all clusters in its identity flow to the DC of its source set.
	\item Correct existing DC-clustering associations from previous time steps based on the marginalisations induced by the identity flow.
\end{enumerate}


How far the procedure can reach back in time to determine a source set of a target cluster needs to be determined through a parametrisation of the method.
This parameter can be considered as the method's history horizon and must be given in number of time steps.
Henceforth, we will refer to a particular configuration of the method as \textit{$x$-step history}, where $x$ specifies the number of snapshots the algorithm reaches back in time.


\section{Consistency in evolutionary clustering}
\label{method:sec:totconsist}
So far, we have focused on the features we consider desirable to define and, ultimately, to detect DCs.
The presented method is a direct implementation of these features, thereby delivering qualitatively satisfying DC structures, when using our custom features as quality criteria.
A more objective approach for quality assessment consists in exploring the auto-correlation of the members between time points.
Note that we consider a DC to be a sequence of sets of clusters, however, the members of a DC at a given time point are all data sources that belong to the clusters within the set of clusters from this time point.
The auto-correlation is given by

\begin{equation}
	C(i) = \frac{|c_{i}\cap c_{i+1}|}{|c_{i} \cup c_{i+1}|},
\end{equation}

where $c_{i}$ and $c_{i+1}$ are the sets of members of the DC, $\mathbf{c}$, at time points $i$, respectively $i+1$, and $|x|$ denotes the cardinality of set $x$.

$C(i)$, also known as the Jaccard index~\citep{jaccard1901distribution}, indicates the fraction of identical members among all members present.
As such, it can be understood as an assessment of membership consistency between neighbouring snapshots.
To alleviate the notation we consider a DC, $\mathbf{c}$, that is not present at snapshot $i$ to have a member set of $c_i = \emptyset$.
We define the \textit{total consistency} of a dynamic clustering as the average auto-correlation.
With the average over all DCs and for each DC over all pairs of neighboring time points the DC exists:

\begin{equation} 
	C_{tot} = \frac{\sum_{\mathbf{c} \in \mathbf{DC}}\sum_{i=0}^{n-1}\frac{|c_{i}\cap c_{i+1}|}{|c_{i} \cup c_{i+1}|}}{\sum_{\mathbf{c} \in \mathbf{DC}}{|\mathbf{c}|} - |\mathbf{DC}|},
\end{equation}

with $n$ the total number of snapshots, $\mathbf{DC}$ the set of all dynamic clusters and $|\mathbf{c}|$ the number of snapshots in which the DC $\mathbf{c}$ exists.

This definition excludes creation and destruction events, i.e.~the DC is only present in one of the two neighbouring snapshots.
It thus indicates the overall consistency of existing structures.

If no external criteria exist that allow the determination of a suitable $x$-step history parameter, we argue that choosing a value that maximises the total consistency score is a sensible choice.
Doing so leads to the most consistent temporal structure within the range of DC structures that result from all possible parametrisations.


\section{Use cases}
We present two exemplary use cases for our method.
In the first we analyse the voting behaviour of US senators over the entire history of the USA.
The US congress is well known to be dominated by a two party system over a long period of its existence, so this first analysis serves also as a test case, in the sense that we expect this two party system to be reflected in the outcome of our method.
In a second case, we analyse the contact structure in a population of free-ranging house mice for which we have individual contact data over several generations.
Here, no clear expectations can be formalized \textit{a priori} and the analysis presented hereafter focuses on an ideal parametrization of our method by maximisation of the total consistency.
A more in-depth analysis of this population of house mice is carried out in a separate study~\citep{liechti2020ff}.

\subsection{US senator votes}
In a first example we analyse the voting behaviour of United States senators over the 116 congresses in the history of the USA.
From the raw voting data, obtained from \textit{voteview.com}~\citep{lewis2018voteview}, we create a sequence of relational datasets of US representatives in the senate, with each congress as a time point in this sequence.
In order to create the relational dataset for a single congress, we follow the approach outlined by~\citet{waugh2009party} that creates connections between representatives if they show similar voting patterns.
More precisely, we define the connection strength between two senators during a congress as the fraction of identical votes among all occasions both were balloting.
Based on these pairwise voting similarities we construct a network for each congress on which we apply a standard modularity optimisation heuristics~\citep{clauset2004finding}, implemented by~\citet{csardi2006igraph}, to determine the community structure.
Since this analysis is based on network data we will refer to clusters as communities and dynamic clusters as dynamic communities for the rest of this paragraph.
For each congress the resulting structure groups senators into \textit{voting-communities}, i.e. according to their voting behaviour.

\begin{figure}[h!]
\centering
\makebox[\textwidth][c]{\includegraphics[width=0.95\textwidth]{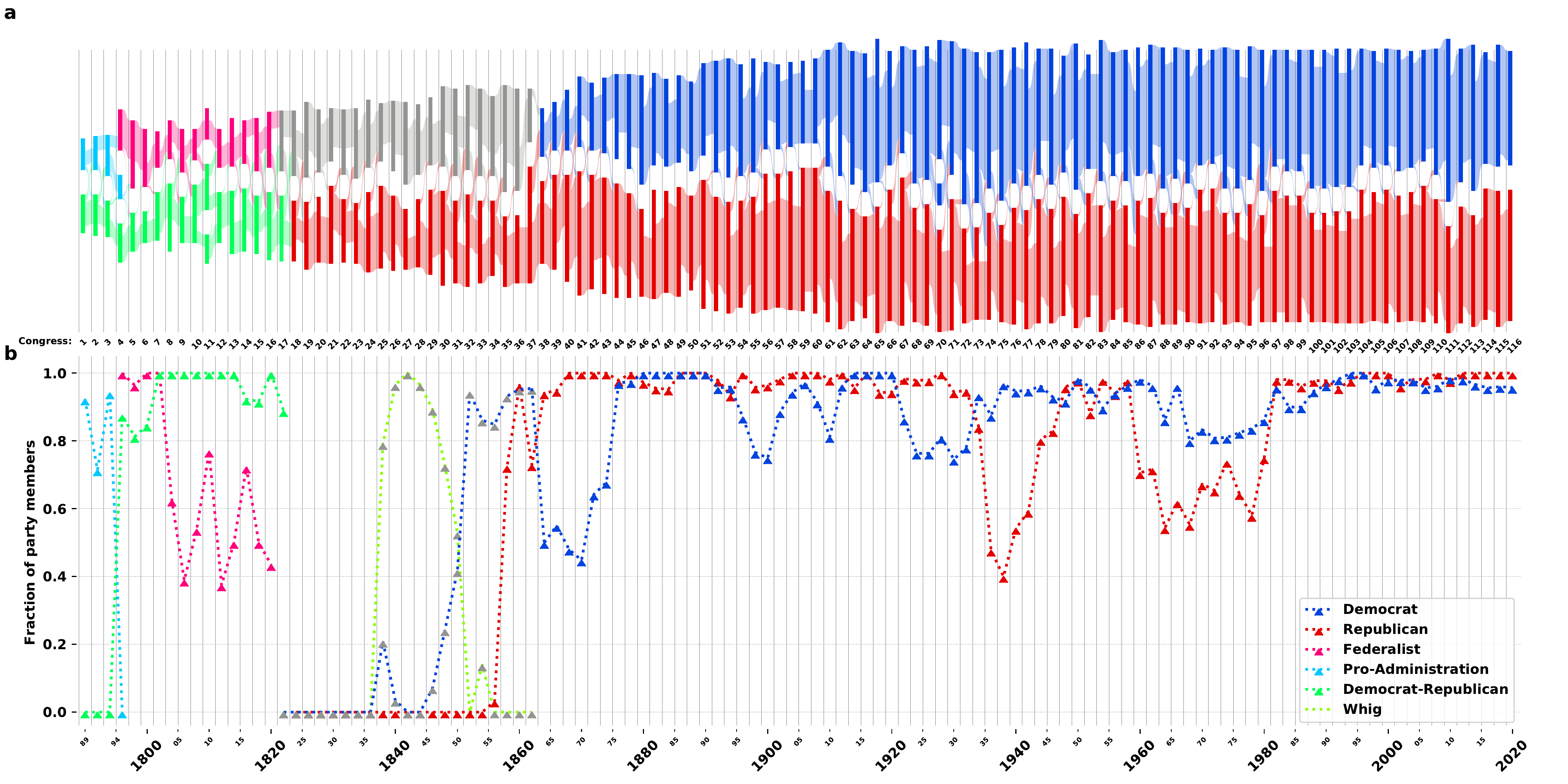}}
\caption{\small%
	\textbf{US senate patterns of similar voting behaviour over its entire history (116 congresses).}
	\textbf{a}, Alluvial diagram illustrating dynamic \textit{voting-communities} (see main text for details) of senators with similar voting behaviour.
	If more than 50\% of all senators belonging to a dynamic voting-community belong to the same party, the dynamic community is coloured with the party colour (see panel b for the corresponding legend).
	If no party holds more than 50\% of the members - the case for a single dynamic community only - then the dynamic community is coloured in gray.
	\textbf{b}, Fraction of community members that belong to the dominant party within the related dynamic voting-community.
	For each of the coloured dynamic voting-communities in panel (a) we report the fraction members at each time point that belong to the colouring party. 
	Dashed lines with gray markers report the fraction of the two best represented parties (Whig and Democrats) in the gray dynamic voting-community that was not dominated by a single party.
  \label{method:fig:uc2}
}
\end{figure}

The political system in the USA is said to be polarised into two fractions~\citep[for an overview see][]{farina2015congressional} and thus we expect a congress to present predominantly two voting-communities.
The alluvial diagram in~\autoref{method:fig:uc2}a indeed shows, with very few exceptions (4th, 11th and 70th congress), consistently two voting communities per congress.
A sequence with community structures as rudimentary as this is an ideal showcase to highlight the majority based association criterion that our method is based on:
It simply allows to identify which of the two voting-clusters in a latter congress relates to which voting-cluster in the former congress (assuming a 1-step history).
Indeed, when applying our method to this sequence, we find that from the 38th congress onward the US senate is essentially divided into the same two dynamic voting-communities.
Also note, that the output is largely parameter-independent:
for values bigger than a 1-step history, the dynamic community structure remains unchanged.
Setting the history parameter to 1-step, simply creates new dynamic communities for those congresses with an additional third voting-community, i.e. for the 4th, 11th and 70th congress.
In~\autoref{method:fig:uc2}a we colour each dynamic voting-community with a party colour, if more than 50\% of the ensemble of member senators belong to the same party (a single dynamic voting-community ends up not being occupied by a party).
The resulting alluvial diagram is a striking illustration of the two party system that has dominated congress in the USA over almost 170 years.
As such, this exemplary use case provides an ideal example to validate or method.
However, we can use the resulting dynamic voting-communities to further explore the dynamic of voting behaviour in the US senate:
We now ask to what extent these dynamic voting-communities are a reflection of the party associations of their senator members.
~\autoref{method:fig:uc2}b reports for each voting-community dominated by a party the fraction of members belonging to the dominant party.
We find that the dynamic community dominated by the Republicans (in red) originated in the 18th congress, which is much before the Republican party came into existence (1850s).
Thus, the Republicans did not found a new voting-community but integrated and took over an existing one.
From congress 73 to 78, the Republican voting-communities become considerably smaller, indicating a shift towards the voting behaviour from the Democratic dynamic voting-cluster.
This shift coincides with the New Deal Coalition.
Further, during the 74th and the 75th congress, the Republicans represented only about 50\% of their dominated voting-community, with the Democrats remaining dominant in theirs.
We conclude that the Republicans were not just less well represented in the senate, but also that some senators from the Democrats were showing a voting behaviour closer related to the one of the Republicans, leading them to belong to the Republican voting-community.

For the Democrats the origin is reversed.
Their dynamic voting-community comes into existence in the 38th congress, towards the end of the Civil War.
Before that, the Democrats belonged to the dynamic voting-community that was not dominated by a single party.
For several congresses it was the Whip party that dominated in the associated voting-communities, only towards the end of its life-span the Democrats became the main party in this dynamic voting-community.
It is interesting to note, that at the end of the Civil War a new voting-community emerged that was initially not dominated by the Democrats, but they became, with time the dominant party.

In more recent history the political system in the USA is said to be highly polarised, potentially more than it ever was throughout its history~\citep{farina2015congressional}.
In terms of dynamic voting communities we note fewer transitions between dynamic voting-communities (fluxes from one colour to the other in~\autoref{method:fig:uc2}a) in the last 20 year, with no transitions at all in the last five congresses.
While this does not exclude the possibility that some senators changed party affiliation, it shows that none of them switched dynamic voting-communities and thus changed his/her voting behaviour drastically.
Regarding the question whether the current polarisation is unique in history, we can only add to consideration that another period with few or no transitions between dynamic voting-communities existed, from the 44th to the 55th congress.

\subsection{Social networks analysis in wild house mice}
As a second example, we analyse the social structure of an individually tagged population of free-ranging house mice, \textit{Mus musculus domesticus}~\citep{konig2012complex,konig2015system,geiger2018longitudinal,ferrari2019fitness}.
In this study, mice of a minimum weight of 18 grams are tagged with an RFID-chip rendering them detectable by antennas situated at the fourty artificial nest boxes distributed within a barn containing the population.
We use concurrent stays of tagged mice in these nest boxes as an indication for social contacts between individuals and create, based on this assumption, contact networks by aggregating the duration of concurrent stays over fixed observation periods.
Following this approach, we get 42 aggregation periods of 14 days of data collection over the total observation period of two years from 2008 to 2010.
Each network in the resulting sequence of networks, also called time-window graphs~\citep{holme2015modern}, is then partitioned using the community detection algorithm by~\citet{rosvall2008maps} in order to obtain a sequence of clusterings.
Finally, we use this sequence of clusterings as input for our algorithm and analyse how different values of the history parameter affect the dynamic clustering structure.
Since this analysis is based on network data we will refer to clusters as communities and dynamic clusters as dynamic communities for the rest of this paragraph.

\begin{figure}[h!]
\centering
\makebox[\textwidth][c]{\includegraphics[width=0.95\textwidth]{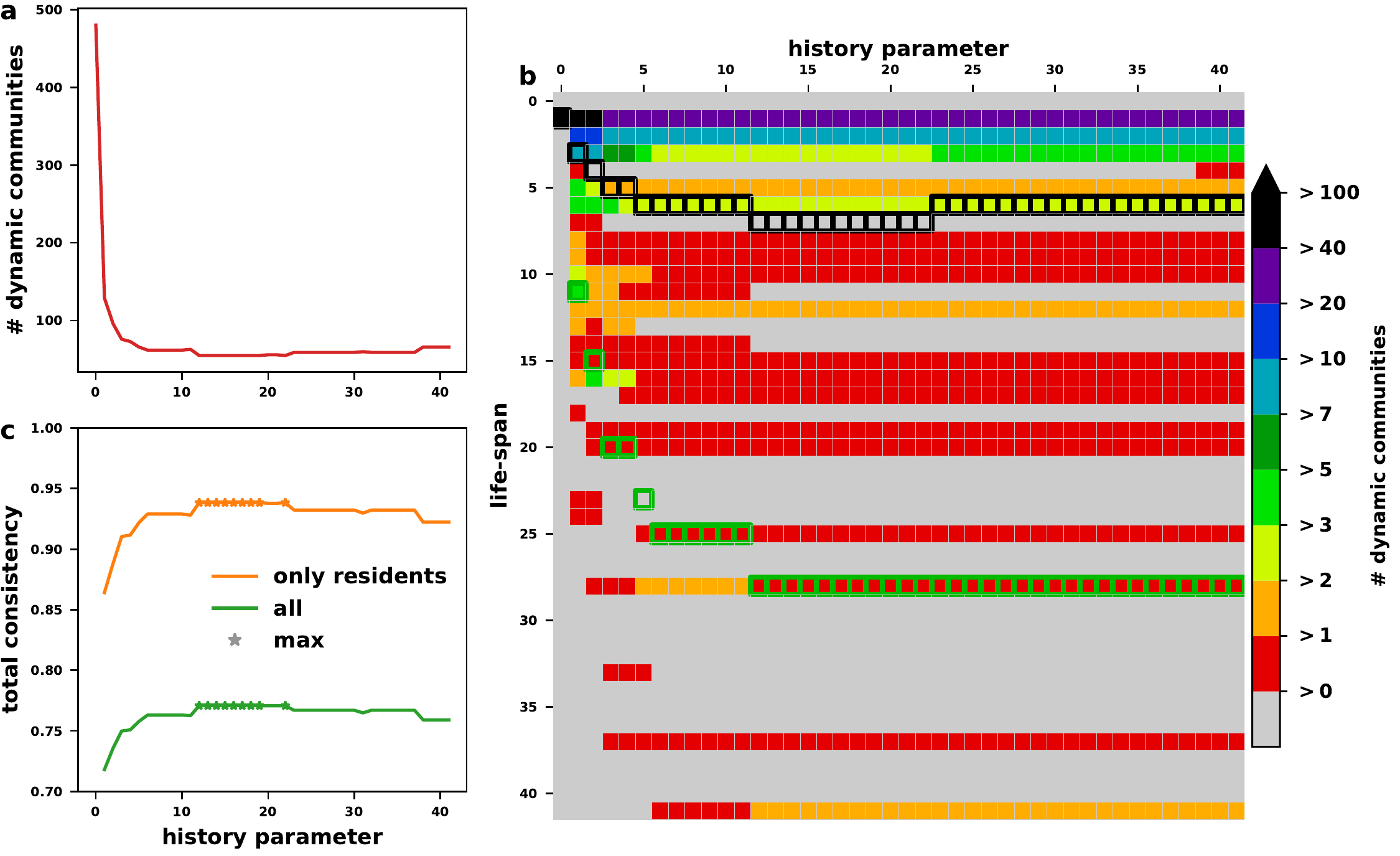}}
\caption{\small%
	\textbf{Characteristics of the dynamic community structure.}
	\textbf{a}, Total number of detected dynamic communities for the range of investigated values of the $x$-step history parameter.
	\textbf{b}, Distribution of life-spans of dynamic communities for various values of the history parameter.
		For each distribution (column) the average is reported by a black border around the bin the average falls into, while a weighted average is reported by green borders.
		The weighted average is to be understood as the life-span of the dynamic community an average house mouse resides in.
		From a 5-step history parameter on, the distribution show high consistency.
		Notable is a single homogeneous discontinuity that spans over 12 time steps, resulting in a remarkable increase of the weighted average.
		It also marks the start of a parameter region, from 12- to 22-step, with maximal total consistency.
	\textbf{c}, Total consistency, for different values of the history parameter.
		Illustrated in green is a version including all individuals to the auto-correlation of two consecutive points, whereas only resident individuals are considered in the orange curve.
		The configurations with maximal total consistency are marked with a star.
  \label{method:fig:uc1}
}
\end{figure}

In~\autoref{method:fig:uc1}a, we show the number of different dynamic communities detected as a function of the history parameter.
While the number of distinct dynamic communities drops drastically within the parameter range from 0 to 5 steps, it remains remarkably stable for higher values.
~\autoref{method:fig:uc1}b provides a more detailed illustration of the same relation by adding the distribution of life-spans of the detected dynamic communities for each value of the history parameter.
Similar to the count, the distributions of life-spans of the detected dynamic communities, also, remain remarkably stable for parameter values bigger than 5 time steps.
We conclude that homogeneous discontinuities extending over more than 5 time steps, i.e. more than 70 days, are rare, with a notable exception of a single decomposition spanning over 12 time steps.
Even though homogeneous discontinuities are mainly short lived, we consistently find (for history parameters bigger than 5 time steps) dynamic communities that persist over a significant fraction of the observation period.
We thus hypothesise that this population of free-ranging house mice is organised in a social structure of rather stable dynamic communities.
We leave an in-depth analysis of the temporal dynamics in the social structure of this population to further studies. 

~\autoref{method:fig:uc1}c presents the total consistency score for the applied range of the history parameter.
It is maximal for a parameter range from $12$ to $22$ steps, which corresponds also to a range of consistent distributions of life-spans (see ~\autoref{method:fig:uc1}b).
Thus, we argue that a 12-step history would be a sensible parametrisation of our method when applied on the community structure within this population of house mice.

\clearpage


\section{Discussion}

We have presented an algorithm that detects dynamic clusters (DCs) in a sequence of snapshots of time stamped data.
It is based on a precisely defined set of features that allow new DCs to form, old ones to disappear, as well as existing DCs to shrink, grow or to transiently split and merge.
The algorithm only depends on a time-series of cluster associations and is, therefore, compatible with any clustering method for non-relational and relational data.  
Thus the user can choose the most suitable clustering method according to the particular study and/or dataset at hand.
Furthermore, this minimal input leads to an efficient scalability of our method with the size of the dataset, i.e.\ the number of data sources present.
It scales linearly with the number of data sources, if the number of clusters detected does not depend on the size of the dataset. 
Consequentially, our method is unlikely to ever figure as the computational bottleneck in the analysis of temporal data.
Only in the limit case when the number of clusters per snapshot scales linearly with the number of data sources is the scalability comparable to the one of a typical clustering method (see~\autoref{scalabilitySI} for further details).

Identifying clusters in a dataset requires an algorithmic procedure that classifies data sources.
Such an algorithm does, in principle, not need to stem from an explicit definition of what a cluster should be.
It can be based on relations between single data points and thus only implicitly define the concept of a cluster - e.g.~two feature vectors need to be closer than a certain distance, in order for their respective data source to belong to the same cluster.
As a result, clustering methods implement a variety of, at least partially, not well defined concepts of a cluster.
This is not only a challenge for traditional clustering, both in relational and non-relational data, but perhaps even more so in data with a temporal resolution.
Therefore, emphasis should be put on a clear outline of the features that define dynamic clusters, be it in the presentation of a new algorithmic procedure, as we do here, or in the application of an existing one.

Finally, this diversity in concepts calls for objective measures that allow to quantify structures deduced by evolutionary clustering methods.
With the total consistency measure defined in this study, we present a measure that does not only allow to objectively parametrise the novel method but also permits to compare different dynamic clusterings quantitatively.

Our method allows to detect transient homogeneous decompositions in dynamic clusters that are present at longer time scales.
Here it is important to clarify, that the detected dynamic clusters differ from the clustering that would result from simply expanding the aggregation period.
This, in part, because transient decompositions that are not homogeneous, i.e.~a dynamic cluster decomposes and some of its parts recombine with (parts of) a different dynamic cluster, might lead to a single cluster including all involved data sources when aggregating over the entire duration of the decomposition.
Another difference is the information that our method retains about transient sub-clusters.
Even if the algorithm determines that, for a given snapshot, several clusters belong to the same DC, the composition of this DC in terms of clusters remains available.
This information on the temporal dynamics within a DC is lost if one simply expands the aggregation period.
Gaining access to the temporal within DC dynamics is the stand alone feature of this novel method and allows to gain insights on within-DC processes, such as the presence and course of sub-clusters or fission-fusion processes~\citep{aureli2008fission}.

  {\small

  }
  \clearpage
  \begin{center}
	  \section*{Supplementary Materials:}
          \addcontentsline{toc}{section}{\protect\numberline{}Supplementary Materials}
  \vspace*{1cm}\textbf{\Large\longTitle}
  \end{center}
	\newcounter{sisection} \setcounter{sisection}{0}
	\newcounter{sisubsection}[sisection] \setcounter{sisubsection}{0}
	 
	\renewcommand{\thesisection}{\textsc{SI Section}\ \arabic{sisection}}
	\renewcommand{\thesisubsection}{\textbf{\arabic{sisection}.\arabic{sisubsection}}}
	 
	\newenvironment{sisection}[1]%
	{
	\refstepcounter{sisection}
	\vspace{1ex}
	\flushleft\makebox[7ex][l]{\textbf{\large SI \arabic{sisection}}} \textbf{\Large #1}\\
	\vspace{1ex}
	}%
	{}
	\newenvironment{sisubsection}[1]%
	{
	\refstepcounter{sisubsection}
	\vspace{1ex}
	\flushleft\makebox[7ex][l]{\textbf{SI \thesisubsection}} \textbf{\large #1}\\
	\vspace{1ex}
	}%
	{}
  \setcounter{equation}{0}
  \setcounter{figure}{0}
  \setcounter{table}{0}
  \setcounter{page}{1}
  \makeatletter
  \renewcommand{\theequation}{S\arabic{equation}}
  \renewcommand{\thefigure}{S\arabic{figure}}
  \renewcommand{\thetable}{S\arabic{table}}
  \renewcommand{\bibnumfmt}[1]{[S#1]}
  \renewcommand{\citenumfont}[1]{S#1}
\sisection{Snapshot representation of data with a temporal resolution}
\addcontentsline{toc}{subsection}{\protect\numberline{}Snapshot representation of data with a temporal resolution}
The presented method for the detection of dynamic clusters requires a time-series of clusterings of data sources.
Time-series data present the particularity that a data source might contribute several data points each of which being associated to a discrete time stamp, thus forming a snapshot of the system associated to this particular time stamp.
For some data this time stamp is given naturally through the process of data collection, for other data the individual data points require binning onto a series of discrete time stamps, before a classification in the form of a clustering can be performed.
Binning consists of aggregating multiple data points from the same data source such that for each snapshot a data source maximally contributes a single data point.

Formally, the raw data for a sequence of snapshots must be present in form of a series of $T$ clusterings of individual data sources $\{G_{i} | i \in \mathbb{N}; i \le T\}$ associated to $T$ time points, $\{t_{i} | i \in \mathbb{N}; i \le T\}$, respectively.
At any time point $t_{i}$ the clustering $G_{i}$ consist of a set of $M_{i}$ clusters, $G_{i} = \{g_{{i}, \alpha} | \alpha \in \mathbb{N}; \alpha \le M_{i}\}$.
For short, we will refer to the time point $t_i$ simply by its index $i$. 

Consider a DC, $\mathbf{c}$, of length $L$ with  $\sigma_\mathbf{c}: \{j \in \mathbb{N}; j \le L_\mathbf{c}\} \to \{i \in \mathbb{N}; i \le T\}$ a mapping of the index from the time-series of length $L$ of member sets of $\mathbf{c}$ to the time-series index of the sequence of snapshots.
The function $\sigma_\mathbf{c}$ is to be understood as the mapping that yields for an index in the time-series of the DC the corresponding index of the time-series of snapshots.
Thus $\sigma_\mathbf{c}$ allows to place an event within the life-span of $\mathbf{c}$ onto the time-series representing the dynamic dataset.
Let $c_i$ be the set of members of $\mathbf{c}$ at the time point $i$, respectively, at the time point $j$, with $\sigma_\mathbf{c}(j)=i$, of the time-series of its member set.
\sisection{The life-cycle of a dynamic cluster}
\addcontentsline{toc}{subsection}{\protect\numberline{}The life-cycle of a dynamic cluster}
\label{lifecycleSI}

We distinguish between six elementary events that might occur in the life-cycle of a dynamic cluster (DC).
These events can be described as follows:

\begin{description}
	\item[\textit{birth}]: The DC $\mathbf{c}$ is born at time point $i+1$ if $\sigma_\mathbf{c}(0) = i+1$ and none of the data sources present in $c_{i+1}$ are members of any other DC at time point $i$.
		Note that while $c_{i+1}$ always exists, the second condition might not always apply, and thus the life-cycle of a DC might not contain a birth event.
	\item[\textit{death}:] A DC dies at ${i+1}$ if $\sigma_\mathbf{c}(L) = i$ and none of its members in $c_i$ are present at time point ${i+1}$.
		Here too, the second condition might not always apply, resulting in an optional presence of a death event in the life-cycle of the DC.
	\item[\textit{growth}:] The size of DC $\mathbf{c}$ at time point $\sigma_\mathbf{c}(j+1)={i+1}$ is given by $|c_{i+1}|$, i.e.\ the number of members present.
		The DC thus grows if there is an increase in its size between $i$ and $i+1$.
	\item[\textit{shrinkage}:] A DC shrinks if its size decreases between $i$ and $i+1$.
	\item[\textit{split}:] The DC $\mathbf{c}$ splits at time point $\sigma_\mathbf{c}(j+1)={i+1}$ if the members of $c_{i}$ are distributed over more than one cluster at time point $i+1$.
		A split thus designates the creation of (at least) one new DC.
		\textit{A priori}, it is not determined whether the splitting DC continues to exist.
		Depending on the applied rules, the DCs holding the remaining members of the split DC at index ${i+1}$ might all be identified as different from the split DC.
		Hence, the split of a DC can also lead to its disappearance.
	\item[\textit{merge}:] Two or more DC merge to become the DC $\mathbf{c}$ at time point $\sigma_\mathbf{c}(j+1) = i+1$, if the members of $c_{i+1}$ are distributed over more than one DC at time point $i$.
		Depending on the specific rules applied the DC at time point ${i+1}$ might be identified as different than, or as one of the involved DC from ${i}$.
		Thus a merge event will lead to the disappearance of at least one DC and might lead to the creation of a new DC at time point ${i+1}$.
\end{description}

\sisection{Evolutionary clustering in a sequence of snapshots representing temporal data}
\addcontentsline{toc}{subsection}{\protect\numberline{}Evolutionary clustering in a sequence of snapshots representing temporal data}
\label{detailsSI}
\sisubsection{Similarity between clusters from consecutive time points}
\label{methodSI:sec:simNeighbouring}

To relate clusters between  time points a measure to quantify cluster similarity is required.
Given that clusters are sets of data sources, that we will call its \textit{members}, a parsimonious choice as basis of a similarity measure between them is the fraction of identical members (\textit{fim}). 

For two clusters, $g_{{i-1}, \alpha}$ and $g_{{i}, \beta}$, from the clustering $G_{i-1}$ at time point $i-1$ and the clustering $G_{i}$ from time point $i$, the \textit{fim} can be defined in three ways:
\begin{align}
	fim(g_{{i-1}, \alpha}, g_{{i}, \beta}) =  \frac{|g_{{i-1}, \alpha}\cap g_{{i},\beta}|}{|g_{{i-1},\alpha} \cup g_{{i}, \beta}|} \label{si:eqt:jaccard}\\
	fim_{\rightarrow}(g_{{i-1}, \alpha}, g_{{i}, \beta}) =  \frac{|g_{{i-1}, \alpha}\cap g_{{i}, \beta}|}{|g_{{i-1}, \alpha}|}\label{si:eqt:mapping}\\
	fim_{\leftarrow}(g_{{i-1}, \alpha}, g_{{i}, \beta}) =  \frac{|g_{{i-1}, \alpha}\cap g_{{i}, \beta}|}{|g_{{i}, \beta}|}\label{si:eqt:tracing},
\end{align}
where $|x|$ denotes the cardinality of set $x$.
The most common definition,~\autoref{si:eqt:jaccard}, is the Jaccard index \citep{jaccard1901distribution}.
It is given by the ratio between the size of the intersection and the size of the union of both clusters.
The Jaccard index is a symmetric measure of cluster similarity.
The two other variations,~\autoref{si:eqt:mapping} and~\autoref{si:eqt:tracing}, are given by asymmetric definitions and use the ratio between the sizes of a single cluster and the union, leading to a time-forward or a time-backward definition of the \textit{fim}.
For two clusters, $g_{{i-1}, \alpha}$ and $g_{{i}, \beta}$, the time-forward definition yields the \textit{fim} of cluster $g_{{i-1}, \alpha}$ present in $g_{{i}, \beta}$, equivalently the time-backward definition, the \textit{fim} of cluster $g_{{i}, \beta}$ present in $g_{{i-1}, \alpha}$.

Independent on the specific definition used, a cluster from time point $i$ might have a non-zero \textit{fim} with several clusters from $i-1$ and vice-versa.
As soon as a non-zero \textit{fim} exists, the biggest \textit{fim}, indicating the best overlap in terms of resident data sources, can be associated to one or several clusters from the neighbouring snapshot.
For clarity, the set of clusters holding the biggest fraction under the asymmetric definition is named \textit{mapping set} in the time-forward case and \textit{tracing set} in the time-backward case.
It follows, that the mapping set of cluster $g_{{i-1}, \alpha}$ consists of the subset from $G_{i}$ that hold the majority of members from $g_{{i-1}, \alpha}$ at time $t_{i}$:
\begin{equation}
ms(g_{{i-1}, \alpha}) = \argmax_{\beta \in \mathbb{N}; \beta \le M_{i}}|g_{{i-1}, \alpha} \cap g_{i, \beta}|.\label{si:eqt:mappingset}
\end{equation}
Equivalently, the tracing set of cluster $g_{{i}, \beta}$ consists of the subset from $G_{i-1}$ that hold the majority of its members at time $t_{i-1}$:
\begin{equation}
	ts(g_{{i}, \beta}) = \argmax_{\alpha \in \mathbb{N}; \alpha \le M_{i-1}}|g_{{i}, \beta} \cap g_{{i-1}, \alpha}|\label{si:eqt:tracingset}.
\end{equation}
In most cases the tracing and mapping set consist of a single cluster.
Only in the case~\autoref{si:eqt:mappingset} or~\autoref{si:eqt:tracingset} yield more than one cluster, i.e.\ the biggest fraction of members is present in several neighbouring clusters, we consider these sets to consist of several clusters.
Note that, whenever we talk about sets containing a single cluster only, our notation might ignore the fact that we are talking about sets, so we might right things like $ts(g_{i,\beta}) = g_{{i-1}, \alpha}$, while technically we should write $ts(g_{i,\beta}) = \{g_{{i-1}, \alpha}\}$.

In addition to the mapping and tracing sets of cluster $g_{{i-1}, \alpha}$, one can define the \textit{tracer set} of cluster $g_{{i-1}, \alpha}$, i.e.\ the set of clusters from $i$ that have a tracing set consisting of only $g_{{i-1}, \alpha}$:
\begin{equation}
	ts_{er}(g_{{i-1},\alpha}) = \{ g_{i, \beta} | ts(g_{i,\beta}) = g_{{i-1}, \alpha} \}\label{si:eqt:tracerset}.
\end{equation}
Equivalently, the \textit{mapper set} of a cluster $g_{{i}, \beta}$ holds all clusters from $i-1$ that have cluster $g_{{i}, \beta}$ as the only member of their mapping set:
\begin{equation}
ms_{er}(g_{{i},\beta}) = \{ g_{{i-1}, \alpha} | ms(g_{{i-1},\alpha}) = g_{{i}, \beta} \}\label{si:eqt:mapperset}.
\end{equation}

Note, the intersection of mapper and tracing set of cluster $g_{{i}, \beta}$ might also be the empty set.

Clusters from neighbouring snapshots might be related through a mapping and/or tracing relation.
In the particular case where the cluster $g_{{i-1}, \alpha}$ form $i-1$ has the cluster $g_{i,\beta}$ as its mapping set and $g_{i, \beta}$ has $g_{{i-1}, \alpha}$ as its tracing set, we will speak of a \textit{bijective majority match} between $g_{{i-1}, \alpha}$ and $g_{i,\beta}$.
A bijective majority match is given, if any of the following conditions is met:
\begin{align}
	ms(g_{{i-1}, \alpha}) = g_{i, \beta}\ \land\ ts(g_{i, \beta}) = g_{{i-1. \alpha}}\label{si:eqt:bijcond}\\
	ms(g_{{i-1}, \alpha}) = g_{i, \beta}\ \land\ g_{{i},\beta} \in ts_{er}(g_{{i-1}, \alpha})\\
	ts(g_{{i}, \beta}) = g_{{i-1}, \alpha}\ \land\ g_{{i-1},\alpha} \in ms_{er}(g_{{i}, \beta})
\end{align}
We consider two clusters forming a bijective majority match, i.e.\ that reciprocally hold their majorities, as different temporal representations of each other.
Note that we also consider the trivial relation of a cluster to itself as a bijective majority map between a cluster and itself.

For any specific definition of the \textit{fim} the normalizing term, i.e.\ the size of the union in the symmetric case and the size of each cluster in the asymmetric cases, can be reduced to resident members only, i.e.\ data sources that contribute a data point to both snapshots.
The definitions of the previously introduced sets readily adapts to a residual-only \textit{fim}, leading to a bijective majority match of residual members as basis for associations of clusters between time points.
From this point on, we only consider resident members, since appearing/disappearing members contribute to a cluster identity in only one of the two time points, and thus provide no information on the similarity of clusters between snapshots.
In doing so we assure that the resulting dynamic clusters become robust against a high turnover rate of members.

\sisubsection{Similarity between clusters from distant time points}

The various sets defined in the previous section describe relations between a single source cluster from one and sets of clusters from a neighbouring snapshot.
To compare subsets from clusterings more than one time step apart we apply the relations described above, provided a minor adaptation:
We extend the source cluster to a source set of clusters.
The above relations will then be applied on each cluster in the set separately.
As a practical example, consider the \textit{mapping set} of $\{g_{{i}, \alpha}, g_{{i}, \gamma}\} \subset G_i$ consisting of the union of both \textit{mapping sets} from $g_{{i}, \alpha}$ and $g_{{i}, \gamma}$:
\begin{equation}
	ms(\{g_{{i}, \alpha}, g_{{i}, \gamma}\}) = ms(g_{i,\alpha}) \cup ms(g_{i,\gamma})
\end{equation}
Using these revised definitions, relations of subsets from adjacent clusterings can be stacked to sequences relating sets of clusters from distant snapshots.
Of particular interest are the recursive application of the mapping set, $ms^n(g_{{i}, \alpha})$, and the recursive application of the tracing set, $ts^n(g_{{i}, \alpha})$.
These two functions will be called \textit{mapping path} and \textit{tracing path} of lengths $n$ and form the basis for a generalization of the bijective majority match of clusters from distant time points.
In particular, if a cluster $g_{i,\alpha}$ has a tracing path that yields a set of clusters from an earlier time point, and this set of clusters has a mapping path that consists of $g_{i, \alpha}$ at time point $i$, we will consider this to be a bijective majority match between $g_{i,\alpha}$ and the earlier set of clusters. 
Thus, a bijective majority match between cluster $g_{i,\alpha}$, from time point $i$, and the set of clusters $ts^n(g_{i,\alpha})$, from time point $i-n$, is given if:
\begin{equation}
	ms^n(ts^n(g_{i,\alpha})) = g_{i, \alpha}\label{si:eqt:bijcondlong}
\end{equation}
Note, for $n=1$ this relation simply reduces to the condition formulated in~\autoref{si:eqt:bijcond} for neighbouring time points: $ms(ts(g_{i,\alpha})) = g_{i, \alpha}$.
Just as in the case of neighbouring time points, we consider the set $ts^n(g_{i,\alpha})$ and $g_{i, \alpha}$ as different temporal representations of each other, if~\autoref{si:eqt:bijcondlong} holds true.
A minor clarification is at hand here:
Since $ts^n(g_{i,\alpha})$ is a set of clusters, $g_{i,\alpha}$ figures as temporally different representation of all of them.
This will be problematic in cases single clusters from $ts^n(g_{i,\alpha})$ were found to belong to different dynamic communities.
Consequentially, we define the \textit{source set} of $g_{i,\alpha}$ as the earliest set of clusters forming a bijective majority match with $g_{i,\alpha}$ while containing only clusters that belong to the same dynamic community.
Recall, that for a dynamic cluster $\mathbf{c}$ its members at time point $i$ are given by $c_i$, so we can define the source set of $g_{i,\alpha}$ as:
\begin{equation}
	ss(g_{i,\alpha}) = \max_{n} \bigg( \bigg\{ ts^n(g_{i,\alpha}) | ms^n(ts^n(g_{i,\alpha})) = g_{i,\alpha}\ \land\ \exists!\mathbf{c}: g_{i-n} \subset c_i; \forall g_{i-n} \in ts^n(g_{i,\alpha}) \bigg\}\bigg).\label{si:eqt:ss}
\end{equation}
Clearly $n < i$, as the earliest time point is the first snapshot.
Since we consider each cluster to hold its own majority, we postulate the trivial relations $ts^0(g_{i,\alpha}) = g_{i,\alpha}$ and $ms^0(g_{i,\alpha}) = g_{i,\alpha}$ and thus $n \ge 0$.
For $n=0$, the conditions within the set defined in~\autoref{si:eqt:ss} will always be met, and thus we can assure the existence of a source set for all clusters.
We add further restrictions by defining a maximal value for the recursion depth of the tracing and mapping paths, $n_{max}$.
This will allow us to set a time horizon beyond which bijective majority matches should no longer be considered.
It follows that the source set, as defined in~\autoref{si:eqt:ss}, depends on this maximal value for $n$ and we need to refine the definition by:
\begin{equation}
	ss(g_{i,\alpha}) = \max_{n;\ n \in [0, \dotsc,  min(i-1, n_{max})]} \big( \big\{ \dotsc \big\}\big).\label{si:eqt:ssrefined}
\end{equation}

Given each cluster has a source set, we can define the \textit{identity flow}, as the ensemble of all clusters that figure in the mapping path and the tracing path between the cluster, $g_{i, \alpha}$ and its source set, $ss(g_{i,\alpha}) = ts^{n^*}(g_{i,\alpha})$:
\begin{equation}
	if(g_{i,\alpha}) = tf \cup mf \circ ss\label{si:eqt:identityflow},
\end{equation}
with the \textit{tracing flow} given by:
\begin{equation}
	tf(g_{i,\alpha}) = \big\{ts^0 \cup ts^1 \cup \dotsc \cup ts^{n^*} \big\}
\end{equation}
and the \textit{mapping flow}:
\begin{equation}
	mf(ss(g_{i,\alpha})) = \big\{ms^0\circ ss \cup ms^{1}\circ ss \cup \dotsc \cup ms^{n^*} \circ ss \big\}.
\end{equation}
To alleviate the notation we omitted $g_{i, \alpha}$ as argument on the right hand sides, i.e.\  we wrote $ts^{n^*}$ instead of $ts^{n^*}(g_{i,\alpha})$, respectively $ms^{n^*} \circ ss$ instead of $ms^{n^*}(ss(g_{i,\alpha}))$.

By definition, the identity flow of a target cluster will contain its source set and the cluster itself.
Any other cluster that might belong to the identity flow takes part in the propagation of the identity of the dynamic community from the source set to the target cluster.
We will thus associate the members of each one of these clusters to this dynamic community at the point of existence of these clusters.

\sisubsection{Marginalisation induced by identity flows}

Before our method is completely specified, another fact requires our attention:
For an identity flow with a length of $n^* > 1$, we might have other identity flows \textit{embedded} into it.
As embedded in an identity flow we consider a sequence of sets of clusters, if the earliest set is tracing back, and the latest is mapping forward to subsets of clusters from the identity flow.
Since we consider all clusters contained in an identity flow to belong to the same dynamic cluster, an embedded sequence is therefore, (i) embedded into a dynamic cluster, and (ii) necessarily shorter than the maximal length of an identity flow, $n_{max}$.
Consequentially, we treat this sequence as a transient part (due to ii) of a homogeneous decomposition (due to i) of the dynamic cluster.
Therefore, we associate the members of all clusters contained within this sequence to the dynamic community embedding it via the identity flow.
We refer to this process as \textit{marginalisation} of the embedded sequence.

Reasoning in terms of identity flows, an identity flow is embedded into another one, and consequentially marginalised, if its source set is tracing back, and its target cluster is mapping forward to the embedding identity flow.
This marginalisation might lead to further identity flows being embed into combinations of the embedding and the now marginalised flow, leading to further marginalisations.
If we follow this logic, carrying out all marginalisations induced by an identity flow becomes an iterative process.

An alternative approach to carry out all marginalisations can be formulated by means of the tracer and mapper sets defined earlier (see~\autoref{si:eqt:mapperset} and~\autoref{si:eqt:tracerset}):
Given the target cluster, $g_{i,\alpha}$, with its identity flow, $if(g_{i,\alpha})$, and its source set, $ss(g_{i,\alpha}) = ts^{n^*}(g_{i, \alpha})$, any other cluster within the sequence of clusterings from time points $i-n^*+1$ to $i-1$ might be subject to a marginalisation if at some stage its tracing path, as well as its mapping path, are included in $if(g_{i,\alpha})$.
We can identify all clusters within the snapshots from $i-n+1$ to $i-1$ that satisfy this condition trough a recursive application of the mapper set starting from $g_{i,\alpha}$:
\begin{equation}
	mt(g_{i,\alpha}) = \big\{ms_{er} \cup {ms_{er}}^2 \cup \dotsc \cup {ms_{er}}^{n^*}\big\},
\end{equation}
which is what we will call the \textit{mapper tree}, and the \textit{tracer tree}, i.e.\ a recursive application of the tracer set, starting from the source set $ss(g_{i,\alpha}) = ts^{n^*}(g_{i, \alpha})$:
\begin{equation}
	tt(ss(g_{i,\alpha})) = \big\{ts_{er} \circ ss \cup {ts_{er}}^2 \circ ss \cup \dotsc \cup {ts_{er}}^{n^*} \circ ss\big\}.
\end{equation}
For any cluster within the mapper tree we know that its mapping path will, at some point, be contained within the identity flow of $g_{i,\alpha}$.
This because its mapping path eventually equals to $g_{i,\alpha}$.
Further, any cluster within the tracer tree of the source set from $g_{i,\alpha}$ will have some stage of its tracing path included in the identity flow, as eventually its tracing path equals to $ss(g_{i, \alpha})$.
It follows, that all clusters being both part of the tracer tree of the source set and the mapper tree of the target cluster have tracing and mapping paths that are included into the identity flow of $g_{i, \alpha}$.
Thus, these clusters are marginalised by the identity flow of $g_{i,\alpha}$.
Formally, all clusters marginalised by the identity flow of $g_{i,\alpha}$ are both part of its mapper tree and the tracer tree of its source set, but not part of its identity flow:
\begin{equation}
	marg(g_{i,\alpha}) = \big\{g\ \big|\ g \in mt(g_{i,\alpha}) \cap tt(ss(g_{i, \alpha})) ; g \notin if(g_{i,\alpha}) \big\}.\label{si:eqt:margens}
\end{equation}

\sisubsection{Definition of a dynamic cluster}
With the definitions of the identity flow of a cluster,~\autoref{si:eqt:identityflow}, and the ensemble of clusters marginalised by it,~\autoref{si:eqt:margens}, we can formulate a definition of a dynamic cluster:

Given a maximal value for the length of an identity flow, $n_{max}$, a dynamic cluster is given by the smallest possible ensemble of clusters that either, (i) have their source set included in the ensemble, (ii) belong to an identity flow of a cluster that has its source set included in the ensemble, or (iii) belong the marginalised clusters of an identify flow from a cluster satisfying (i) or (ii).

The ensemble needs to be as small as possible, as otherwise combinations of ensembles that fulfill this definition still qualify as dynamic communities, eventually leading to a giant dynamic community that includes all clusters present in the dataset.

\sisubsection{Algorithmic application procedure}
For a sequence of snapshots, we will apply the definition of a dynamic cluster iteratively, starting from the first snapshot, consistently ignoring data from all further snapshot.

The maximal value of the length of an identity flow, $n_{max}$, sets a limit for the length over which the identity of a dynamic cluster can span, and consequentially, limits the maximal length of sequences that can be marginalised.
In an iterative application, this values determines how far an identity flow can maximally reach back, and thus, can be understood as some sort of horizon for the history we can modify and take into account.
Therefore, we name this parameter the \textit{history parameter} and will refer to an application of our method with a specific value for it as ``using a \textit{x-step history}''.

In~\autoref{methodSI:tbl:algoDesc} we illustrate the algorithmic procedure.
First, we demonstrate the base case, i.e.\ the application of our definition of a dynamic cluster considering only the first snapshot of the time series.
Second, we walk through a step case, i.e.\ the application of our definition to an arbitrary time point, $i+1$, assuming the method was applied to all previous time points.
While we describe the single steps of the algorithmic procedure, we leave it to the reader to verify that the resulting dynamic clusters indeed satisfy the definition provided above.
%

\clearpage
\begin{center}
	\newcounter{algStepC}
	\newcommand{\algStepName}{stage}
	\newcommand{\theAlgStep}{\algStepName{} \thealgStepC}
	\newcommand{\nextAlgStep}{\theAlgStep{}\refstepcounter{algStepC}}
	\newcommand{\algStep}{\parbox[t]{2mm}{\rotatebox[origin=c]{90}{\textbf{\nextAlgStep}}}}
 \setstretch{1.0}  
 \footnotesize
  \begin{longtable}{@{} m{0.1cm} c m{0.3\textwidth} m{.3\textwidth} @{}}
  \caption{\small
	  \textbf{Step-by-step description of the novel evolutionary clustering method illustrated by means of an example using a 3-step history.}
	For more information on the applied terminology see~\autoref{detailsSI}.
  }\label{methodSI:tbl:algoDesc}
  \\
    & Illustration & Description & Comment \\
  \midrule
  \endfirsthead
  \multicolumn{4}{r}{{\tablename\ \thetable{} -- \textit{continued from previous page}}}
  \\\\
    & Illustration & Description & Comment \\
  \midrule
  \endhead

  \hline \multicolumn{4}{r}{{Continued on next page}} \\ \hline
  \endfoot
  \hline \hline
  \endlastfoot
  \\
  \multicolumn{4}{l}{{\textbf{Base Case:} Cluster-DC association at the first step of the time-series.}}
  \\
    &
  \hspace*{-1cm}
    \begin{minipage}{.4\textwidth}
      \includegraphics[width=\linewidth]{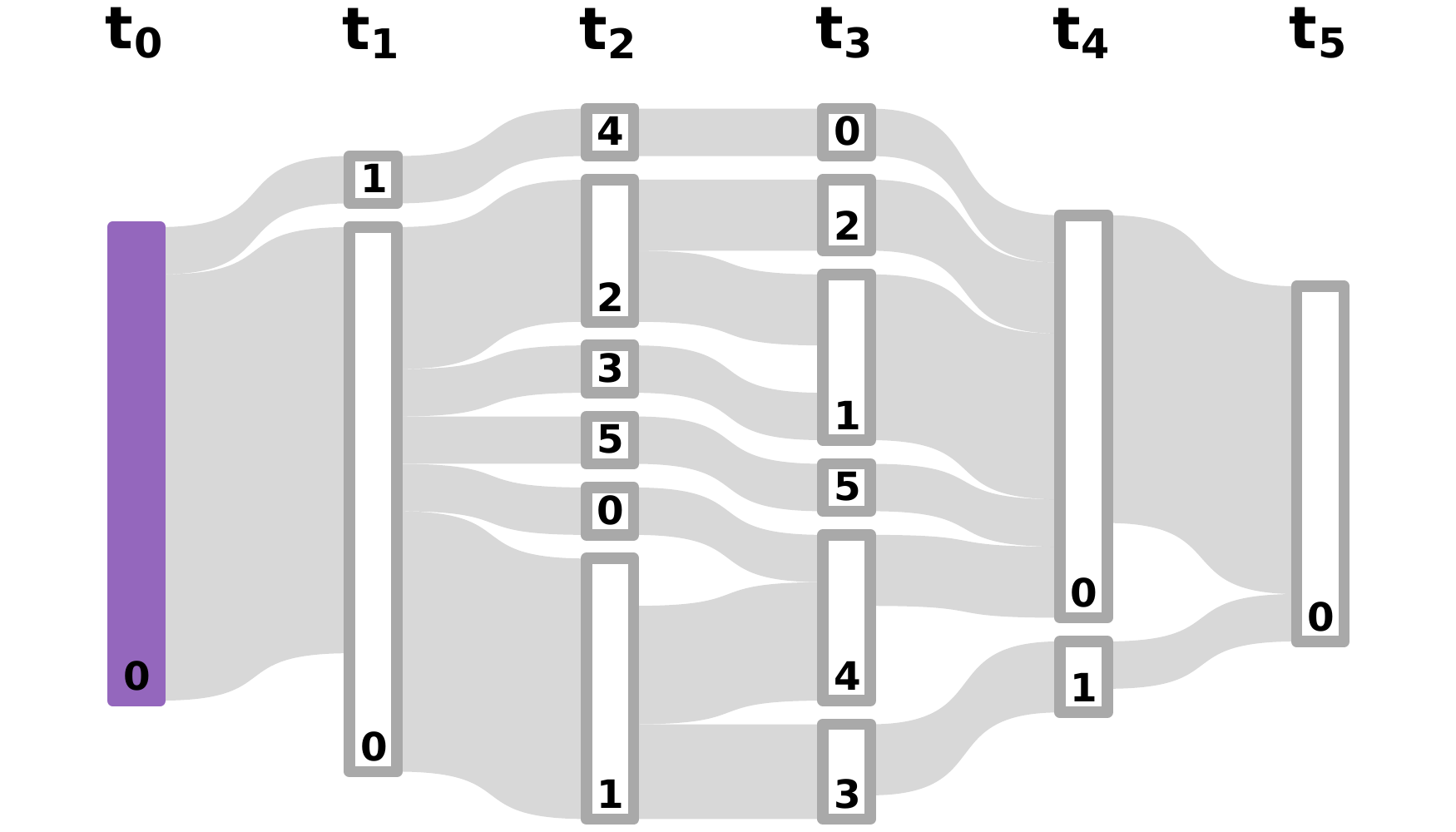}
    \end{minipage}
    &
    At the start of the time-series, $t_{0}$, all present clusters are associated to newly created, distinct DCs.
    &
    In the present example cluster $g_{0,0}$ is the only element in the initial clustering $G_0$.
    It is associated to a new DC indicated in purple.
    \\
  \midrule
  \\
  \multicolumn{4}{l}{{\textbf{Step Case:} Cluster-DC association at time point $t_{i+1}$, given all cluster-DC associations of time points  $t_{i+j}, j \le 0$.}}
  \\
    \algStep
    &
    \begin{minipage}{.4\textwidth}
      \hspace*{0.025cm}
      \includegraphics[width=1.0\linewidth]{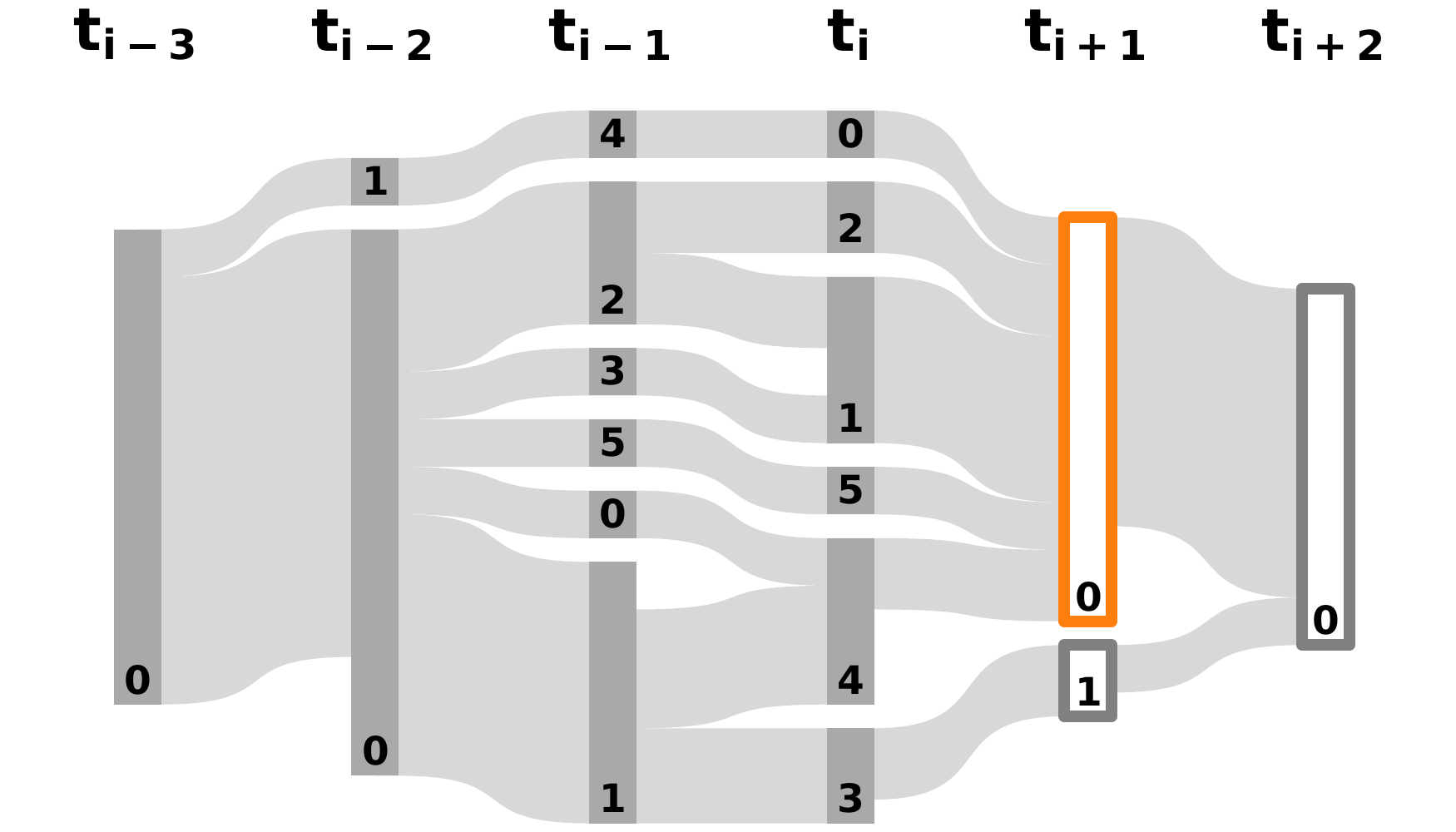}
    \end{minipage}
    &
    Select a target cluster from time step, $t_{i+1}$, that has not yet been associated to a DC.
    &
    The target cluster $g_{i+1, 0}$ is marked in orange.

    The order at which clusters from the next snapshot are processed has no effect on the output of the algorithm.
    \\
    \algStep
    &
    \begin{minipage}{.4\textwidth}
      \includegraphics[width=\linewidth]{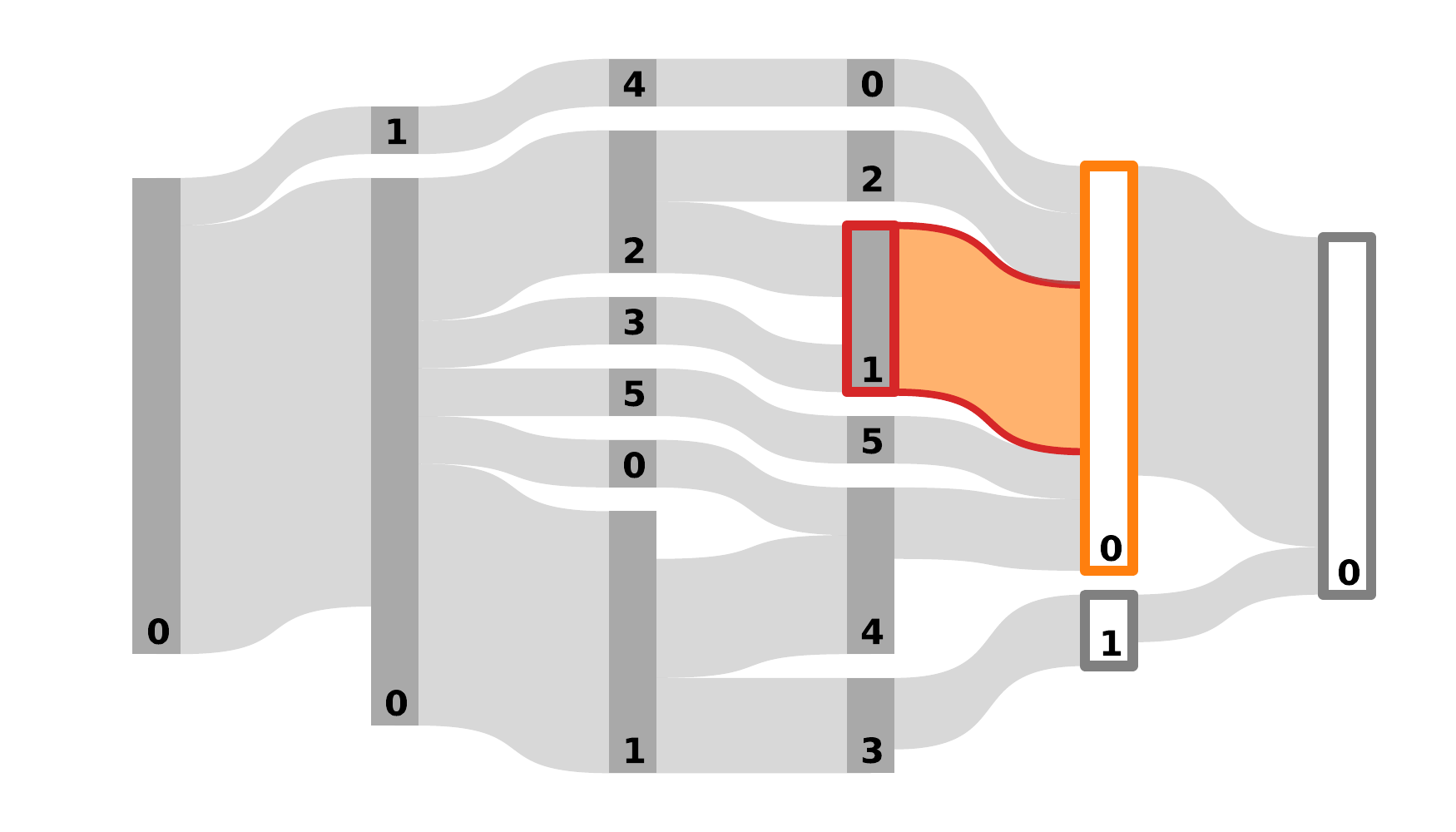}
    \end{minipage}
    &
    Determine the tracing set of the target cluster.
    If the mapping set of the target cluster's tracing set is the target cluster, a tracing flow can be constructed.
    If not, the target cluster is associated to a newly created DC.
    & 
    The tracing flow is colored in orange.
    The tracing set $\{g_{i, 1}\}$ is marked red, its forward mapping flow is marked in red.

    The tracing set holds the majority of members of the target cluster at the previous snapshot, $t_{i}$.
    \\
    \algStep
    &
    \begin{minipage}{.4\textwidth}
      \includegraphics[width=\linewidth]{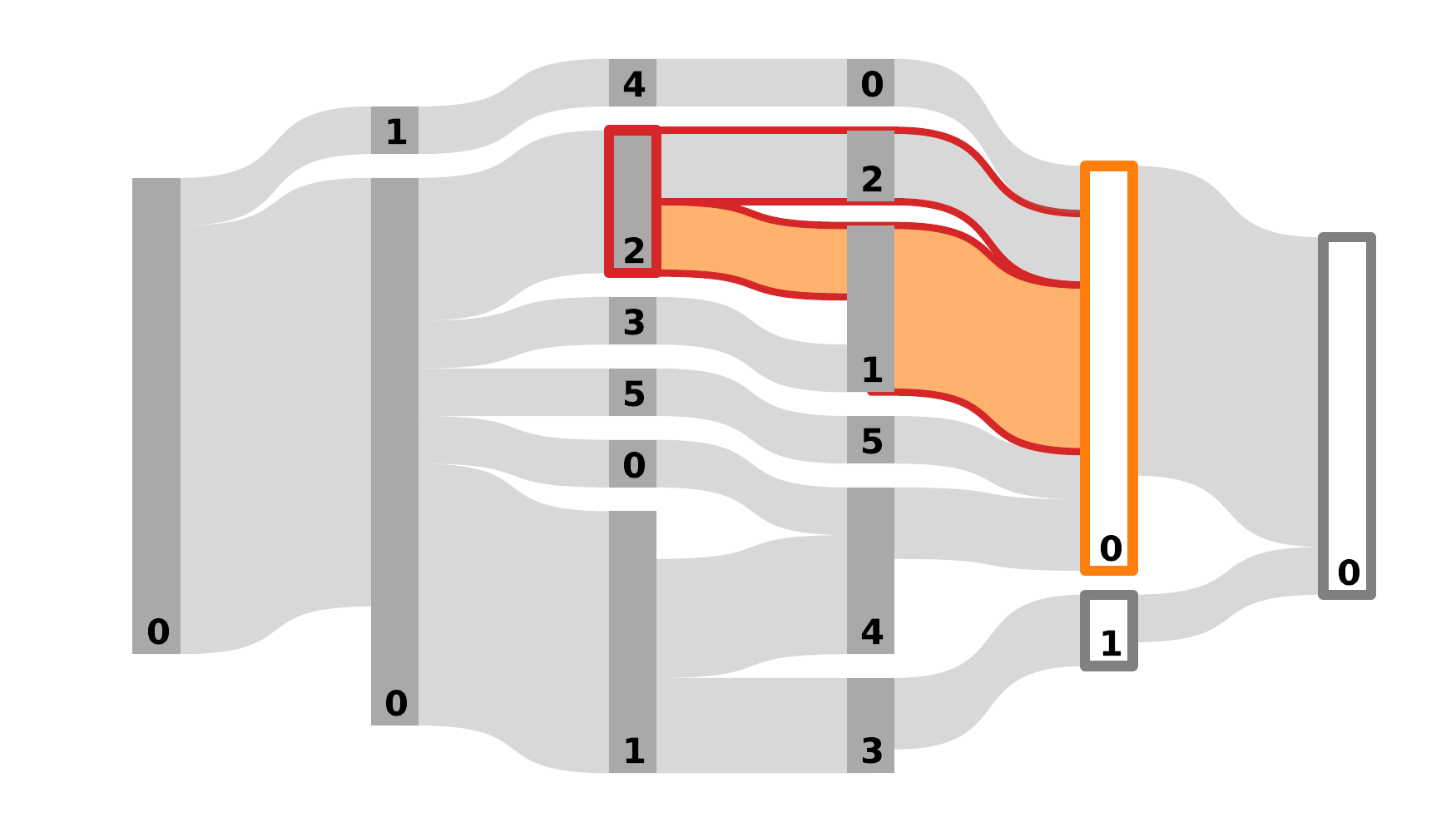}
    \end{minipage}
    &
    Iteratively construct the tracing flow:
    At each iteration identify the next tracing set of the last set in the tracing flow.
    If the next tracing set has a mapping flow that is at some time point contained in the tracing flow, it is added to the tracing flow.
    & 
    The tracing flow is colored in orange.
    The next tracing set is highlighted in red and its mapping flow is marked in red. 

    The tracing flow follows the (not yet known) DC identity of the target cluster back in time.
    Equivalently, the mapping flow follows the DC identity of the last tracing set forward in time.
    \\
    \algStep
    &
    \begin{minipage}{.4\textwidth}
      \hspace*{0.025cm}
      \includegraphics[width=\linewidth]{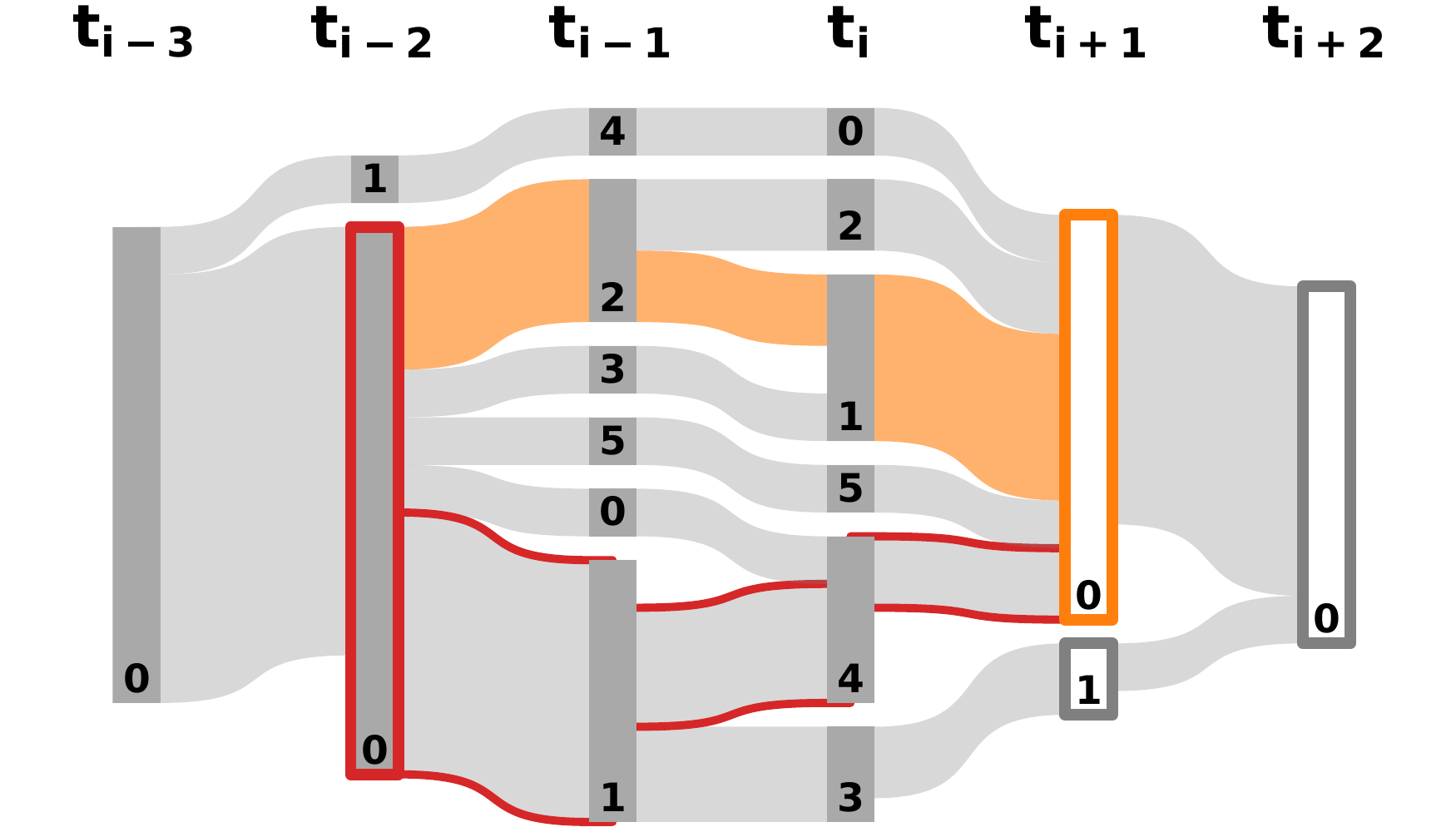}
    \end{minipage}
    &
    The tracing flow ends either if no further tracing set exists, at the defined history-limit or the first time point in the time-series, $t_0$.
    & 
    For the coloring see previous \algStepName{}.

    In the present example a 3-step history was used, thus the tracing flow stops at $t_{i-2}$.
    \\
    \algStep
    &
    \begin{minipage}{.4\textwidth}
      \includegraphics[width=\linewidth]{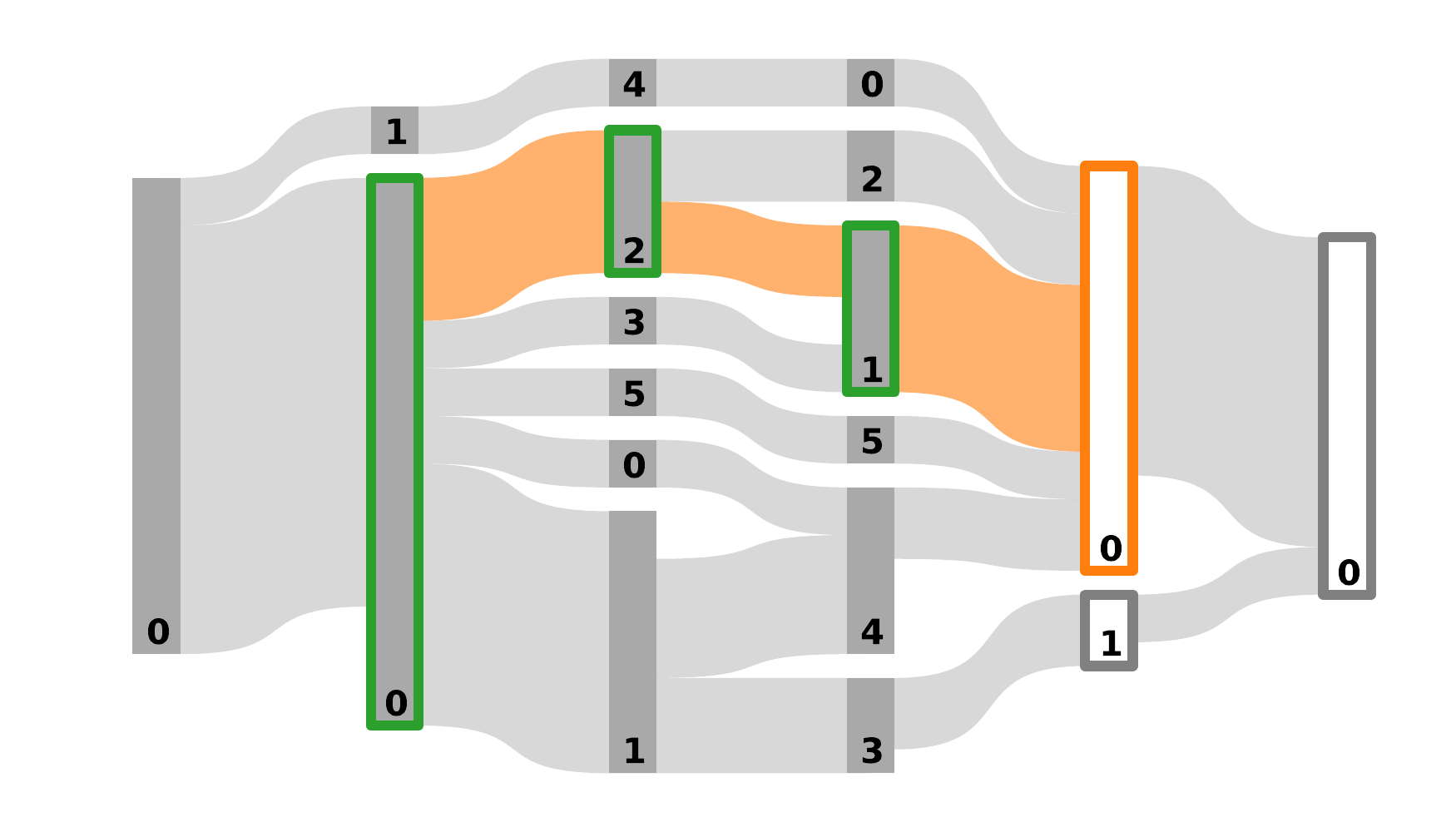}
    \end{minipage}
    &
    Identify all the potential source sets along the tracing flow.
    & 
    The tracing flow is colored in orange with all potential source sets highlighted in green.
    \\
    \algStep
    &
    \begin{minipage}{.4\textwidth}
      \includegraphics[width=\linewidth]{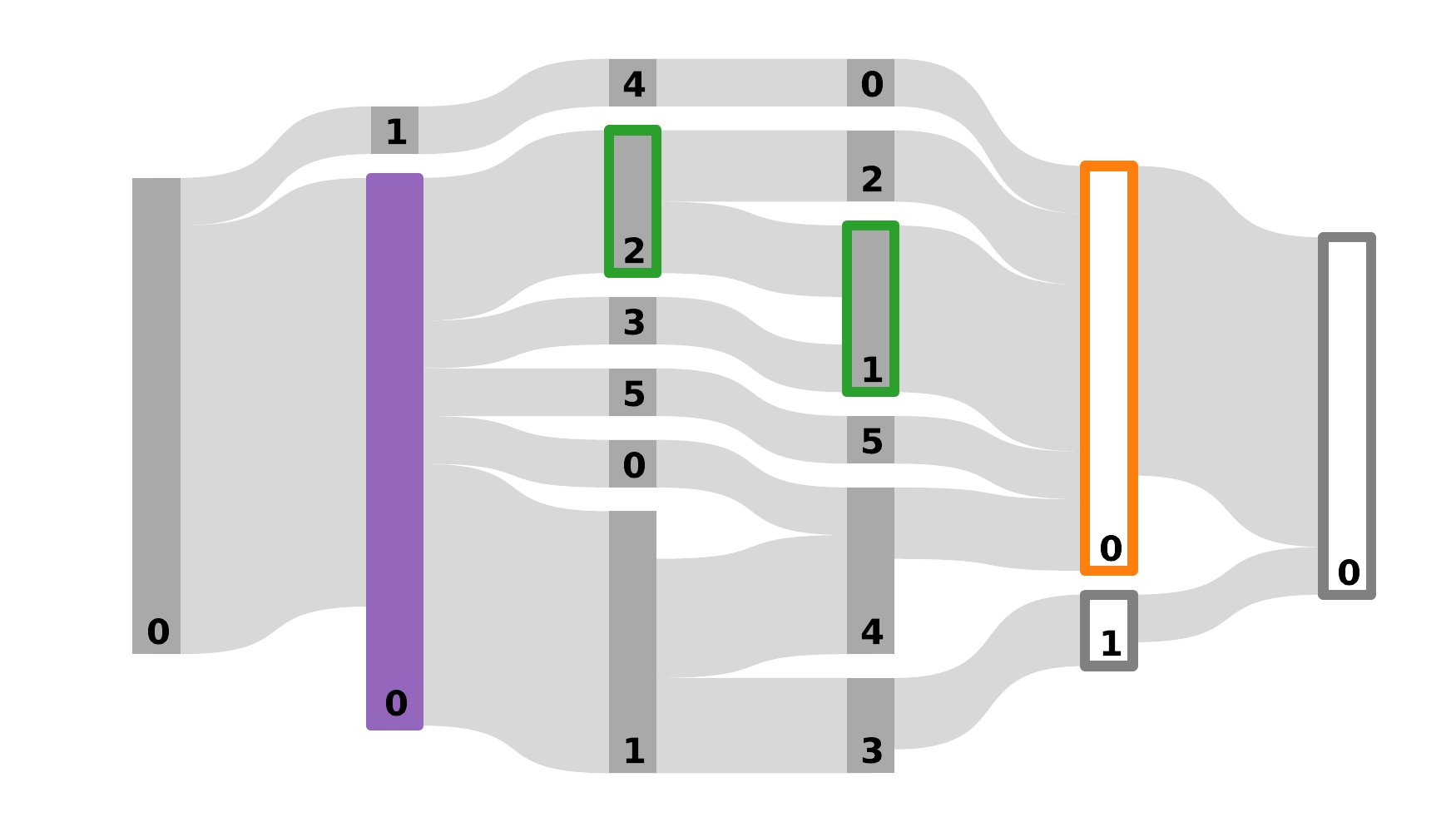}
    \end{minipage}
    & 
    Determine the source set:
    Start from the earliest potential source set and check if the groups in the source set belong to a single DC.
    If no potential source set satisfies the condition create a new DC identity for the target cluster.
    & 
    The earliest candidate source set, $\{g_{{i-2}, 0}\}$, colored in purple, consists of a single cluster.
    The other potential source sets are highlighted in green.
    \\
    \algStep
    &
    \begin{minipage}{.4\textwidth}
      \includegraphics[width=\linewidth]{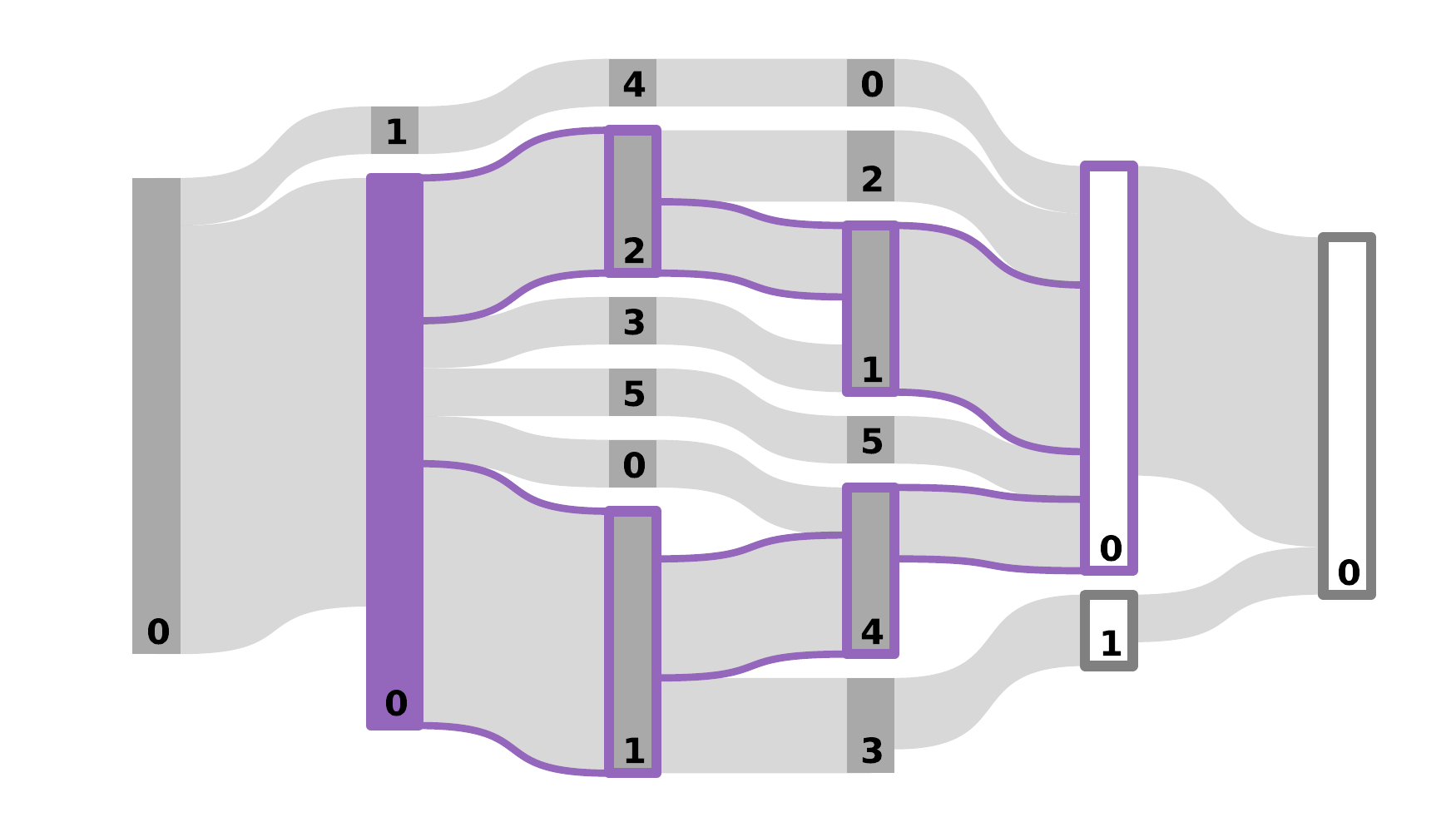}
    \end{minipage}
    &
    Identify the identity flow, i.e.\ all clusters from the tracing flow of the target cluster and the mapping flow of the source set.
    & 
    The source set is colored in purple.
    Mapping and tracing flows with its associated clusters are highlighted in purple.

    Clusters within the tracing or/and the mapping flow propagate community identities in one/two directions.
    \\
    \algStep
    &
    \begin{minipage}{.4\textwidth}
      \includegraphics[width=\linewidth]{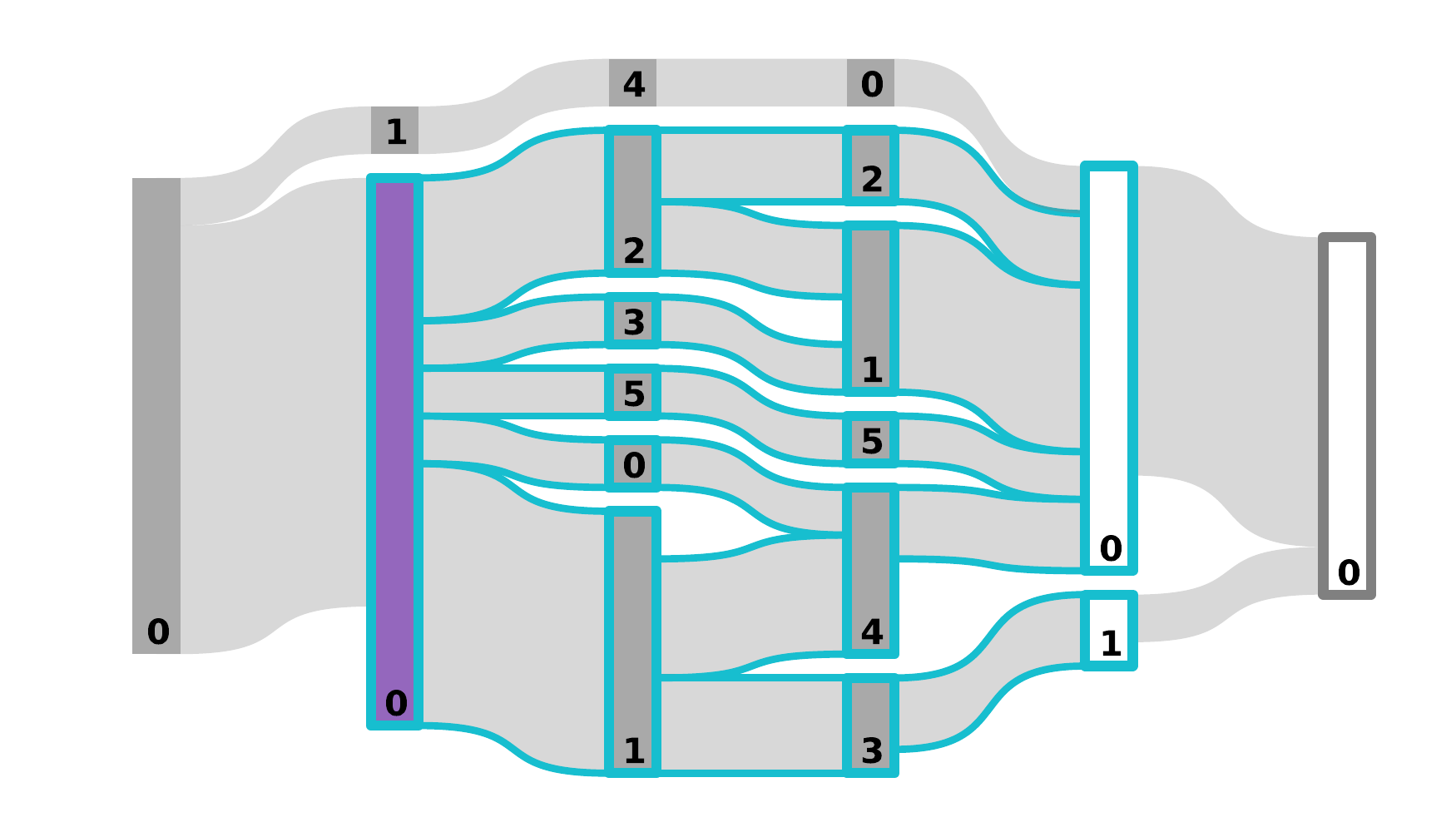}
    \end{minipage}
    &
    Iteratively determine the tracer flow of the source set up to $t_{i+1}$.
    The next set in the tracer flow is given by the tracer set of the previous set, starting with the source set.
    & 
    The tracer flow is highlighted in blue with its source set colored in purple.
    It contains by construction the identity flow (see~\algStepName{} 6).
    The tracer set of a given set consists of all clusters that have the given set as their tracing set.
    \\
    \algStep
    &
    \begin{minipage}{.4\textwidth}
      \hspace*{0.025cm}
      \includegraphics[width=\linewidth]{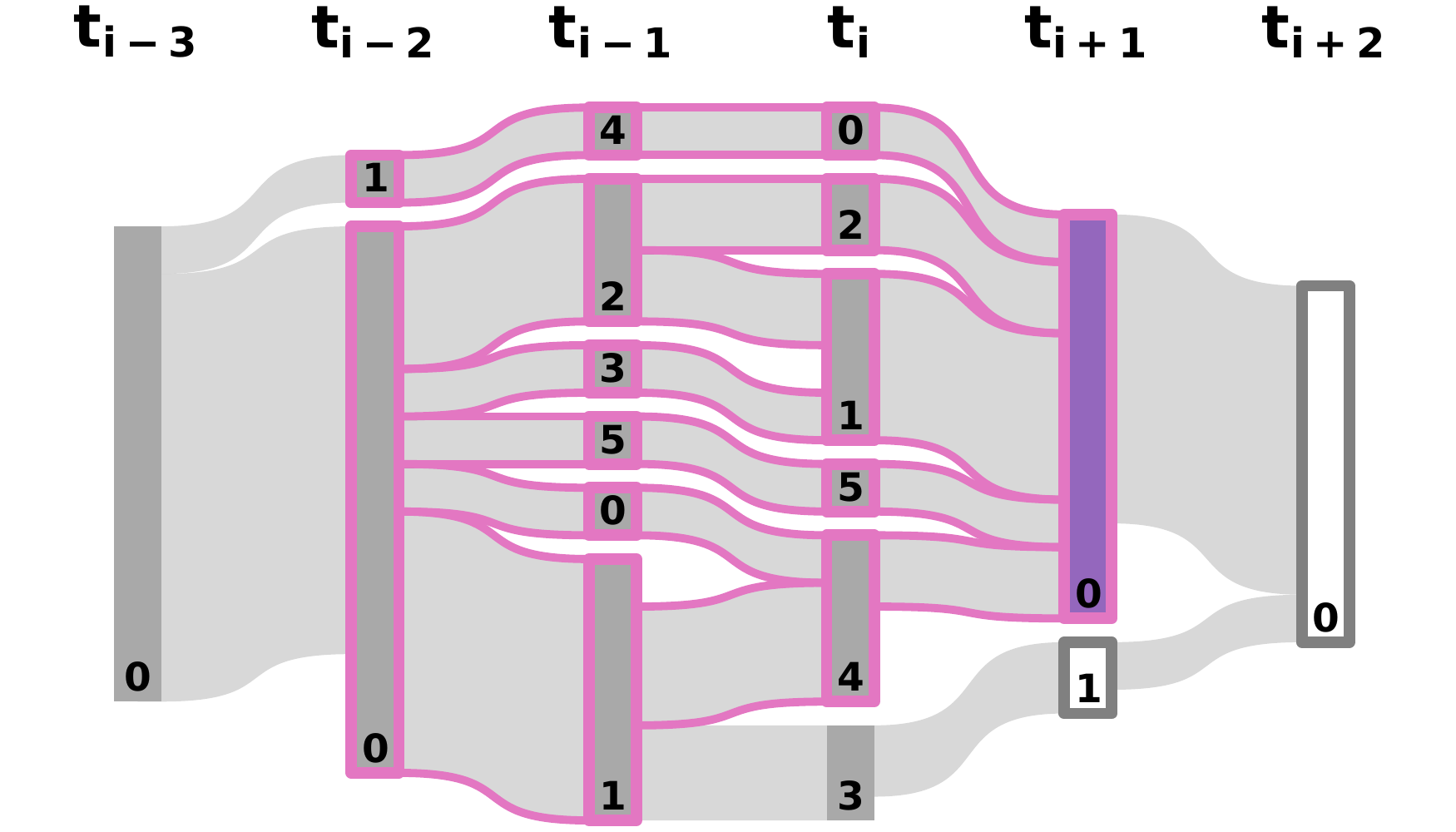}
    \end{minipage}
    &
    Iteratively determine the mapper flow of the target cluster back to the time point of the source set.
    The next set in the mapper flow is given by the mapper set of the previous set, starting with a set containing only the target cluster.
    & 
    The mapper flow is highlighted in pink with its initial set consisting of the target cluster colored in purple.
    It contains by construction the identity flow (see~\algStepName{} 6).
    The mapper set of a given set consists of all clusters that have the given set as their mapping set.
    \\
    \algStep
    &
    \begin{minipage}{.4\textwidth}
      \includegraphics[width=\linewidth]{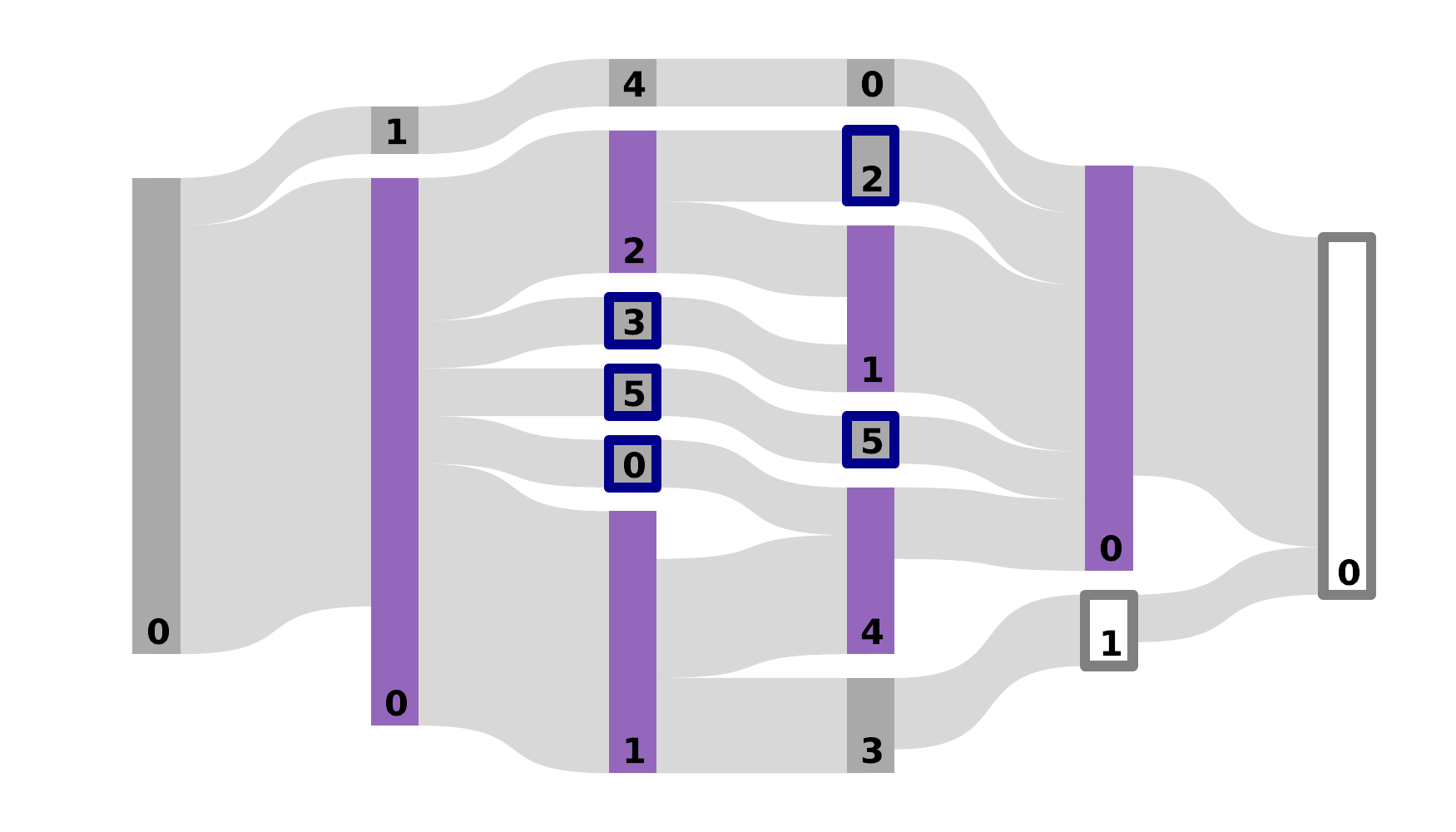}
    \end{minipage}
    &
    At each time point, compute the intersection between the sets from the tracer (\algStepName{} 7), and the mapper (\algStepName{} 8) flow.
    Determine the marginal clusters, i.e.\ clusters in the intersection that do not belong to the identity flow (from \algStepName{} 6).
    & 
    Clusters from the mapping or tracing flow are colored in purple.
    Marginal clusters are highlighted in dark-blue.
    Marginal clusters, if present, do not contribute to the persistence of a DC as they propagate the DC identity maximally in one direction.
    \\
    \algStep
    &
    \begin{minipage}{.4\textwidth}
      \includegraphics[width=\linewidth]{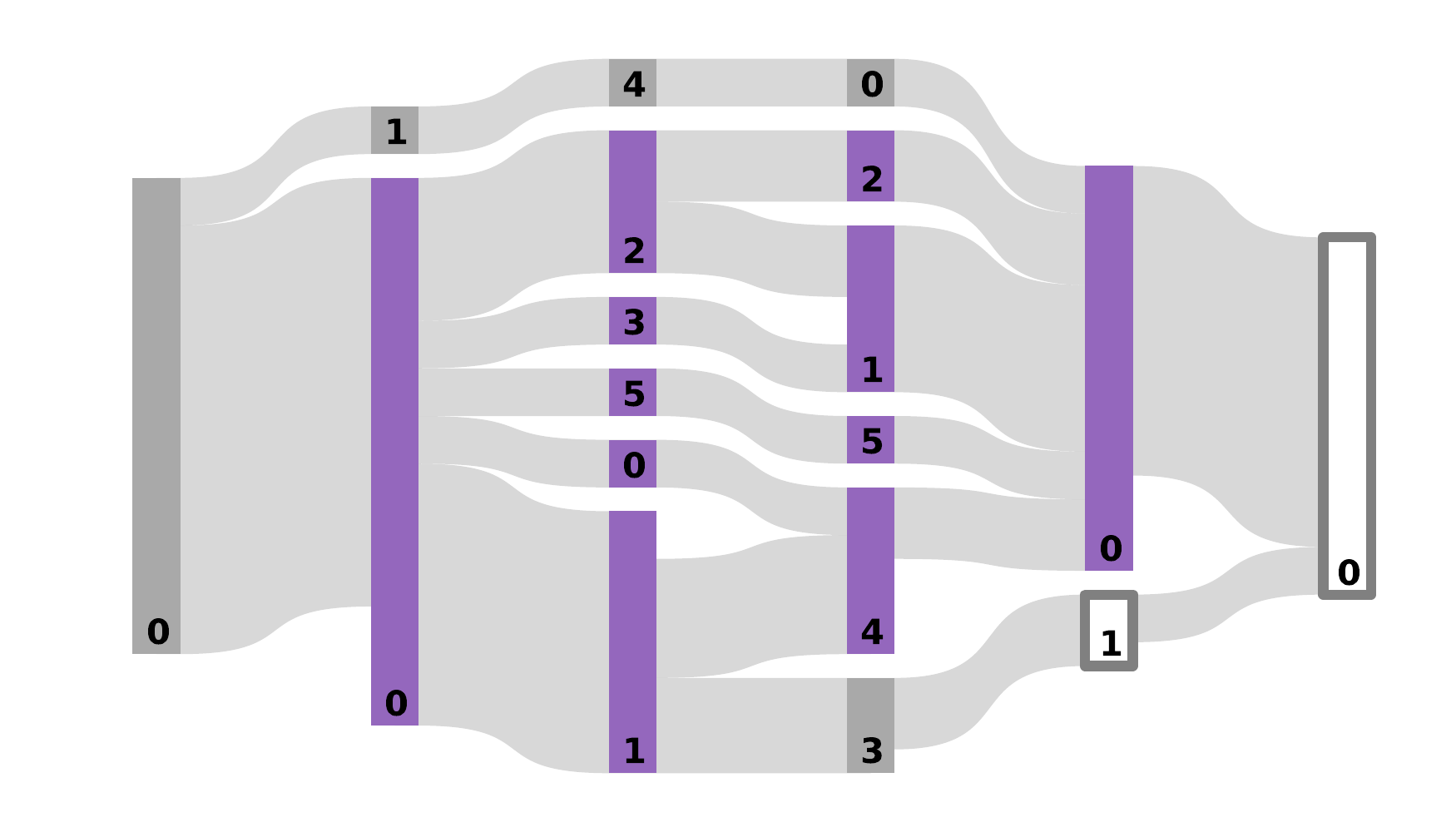}
    \end{minipage}
    &
    Associate both the identity clusters (from \algStepName{} 6) and the marginal clusters (from \algStepName{} 9) to the DC of the source set.
    & 
    The algorithm has successfully associated cluster $g_{i+1, 0}$ to a DC and implemented corrections to the DC structure based on this new association.
    
  \end{longtable}
  \end{center}

\sisection{Scalability of the algorithm}
\addcontentsline{toc}{subsection}{\protect\numberline{}Scalability of the algorithm}
\label{scalabilitySI}
The factors affecting the scalability of our method are: $T$, the number of snapshots in the sequence; $N$, the number of data sources present; $G$, the average number of cluster per snapshot; and $x$ (or $n_{max}$), the history parameter.

The method can be separated into two operational steps:
Firstly, the elementary majority relations, i.e.\ between the clusters from two neighbouring snapshots, are established.
Secondly, the dynamic clusters are defined, which includes the identification of majority relations between sets of clusters from distant time points and the determination of marginal flows.

Elementary majorities result from the similarity between any pair clusters from two neighbouring snapshots, which is an operation $\mathcal{O}(N + G^2)$, scaling linearly with the number of data sources present, $N$, and the number of all possible pairwise combinations of clusters, $G^2$.
The majority set is then determined by the maximal similarity, which for each cluster is an operation scaling with the number of clusters in the neighbouring snapshot.
Therefore, the majority determination from the similarities scales, again, with the number of all pairwise combinations of neighbouring clusters.
Thus, if the number of clusters in a snapshot does not scale with $N$, the determination of elementary majority relations scales linearly with $N$.
In the extreme case, where the number of clusters scales linearly with $N$, calculating elementary majority relations scales quadratically with the number of clusters and thus with $N^2$. 

In a sequence of snapshots of length $T$, $T-1$ majority relations between neighbouring snapshots need to be computed and thus the determination of elementary majority relations grows linearly with the length of the sequence.

Once the elementary majority relations are determined, the algorithm establishes majority relations between distant sets of clusters with a maximal distance given by the history parameter, $x$. At worst, for each cluster, the algorithm finds a majority match reaching $x$ steps back in time, resulting in $x(x+1)$ operations.
This because the procedure first reaches one step back in time and returns to the current position, then two steps and so on, up to $x$ steps back.
With the identity flow at hand, i.e.\ the majority relations between distant time points, the marginal clusters must be determined.
A cluster has an average size of $\frac{N}{G}$ and thus has non-zero similarities with at most $\mathcal{S} = \min(G, \frac{N}{G})$ clusters from a neighbouring snapshot.
If the number of cluster scales linearly with $N$ and if the number of clusters remains roughly constant with $N$, then this term remains roughly constant.
To determine the marginal flows, we need to follow all mapper relations back starting from the target cluster and all tracer relations forward starting from the source cluster set, which both are, in the worst case, operations $\mathcal{S}^{\mathcal{O}(x)}$ and thus scale exponentially with the history parameter.

The establishment of majority relations from distant time points does, for a single cluster, not scale with the length of the sequence, however, the number of clusters scales with $\mathcal{O}(TG)$, and thus is linear in $T$.

To recap, the algorithm scales at worst: linear in the number of snapshots $T$ and the number of data sources $N$, quadratically in the average number of cluster per snapshot $G$ and exponentially with the history parameter $x$.
As a consequence, the method scales linearly in the number of data sources present $N$, if the number of clusters remains roughly constant and quadratically, if the number of clusters scales linearly with $N$. 
Given that clustering algorithms typically relate on pairwise similarity measures between the data sources and thus scale $\mathcal{O}(N^2)$, there is little risk that our method ever appears as the computational bottle neck.
Only in the limit case, when the number of clusters per snapshot scales linearly with the number of data sources, the scaling of our method with the number of data sources will be comparable to the scaling of the clustering method providing the sequence of clusters.

\sisection{Software}
\addcontentsline{toc}{subsection}{\protect\numberline{}Software}
\sisubsection{MajorTrack}
\label{software:MajorTrack}
Evolutionary clustering method, tracking persistent but transiently discontinuous clusters in a sequence of snapshots for temporal non-relations or relational data.

Available on GitHub: \href{https://github.com/j-i-l/MajorTrack}{https://github.com/j-i-l/MajorTrack}

\sisubsection{pyAlluv}
\label{software:pyAlluv}
Package for the visualization of Alluvial diagrams in python.
It is based on core packages from matplotlib~\citep{hunter2007matplotlib}.

Most of the figures in this study were generated using this package.

Available on GitHub: \href{https://github.com/j-i-l/pyAlluv}{https://github.com/j-i-l/pyAlluv}



\begin{thebibliography}{27}
\providecommand{\natexlab}[1]{#1}

\bibitem[{Aureli \emph{et~al.}(2008)Aureli, Schaffner, Boesch
  \emph{et~al.}}]{aureli2008fission}
Aureli, F., C.~M. Schaffner, C.~Boesch \emph{et~al.} 2008.
\newblock Fission-fusion dynamics: new research frameworks.
\newblock \emph{Current Anthropology} \textbf{49}(4):627--654.

\bibitem[{Backstrom \emph{et~al.}(2006)Backstrom, Huttenlocher, Kleinberg
  \emph{et~al.}}]{backstrom2006group}
Backstrom, L., D.~Huttenlocher, J.~Kleinberg \emph{et~al.} 2006.
\newblock Group formation in large social networks: membership, growth, and
  evolution.
\newblock In: \emph{Proceedings of the 12th ACM SIGKDD international conference
  on Knowledge discovery and data mining}, pages 44--54. ACM.

\bibitem[{Chakrabarti \emph{et~al.}(2006)Chakrabarti, Kumar and
  Tomkins}]{chakrabarti2006evolutionary}
Chakrabarti, D., R.~Kumar and A.~Tomkins. 2006.
\newblock Evolutionary clustering.
\newblock In: \emph{Proceedings of the 12th ACM SIGKDD international conference
  on Knowledge discovery and data mining}, pages 554--560. ACM.

\bibitem[{Clauset \emph{et~al.}(2004)Clauset, Newman and
  Moore}]{clauset2004finding}
Clauset, A., M.~E. Newman and C.~Moore. 2004.
\newblock Finding community structure in very large networks.
\newblock \emph{Physical review E} \textbf{70}(6):066111.

\bibitem[{Csardi \emph{et~al.}(2006)Csardi, Nepusz
  \emph{et~al.}}]{csardi2006igraph}
Csardi, G., T.~Nepusz \emph{et~al.} 2006.
\newblock The igraph software package for complex network research.
\newblock \emph{InterJournal, complex systems} \textbf{1695}(5):1--9.

\bibitem[{Dakiche \emph{et~al.}(2019)Dakiche, Tayeb, Slimani
  \emph{et~al.}}]{dakiche2019tracking}
Dakiche, N., F.~B.-S. Tayeb, Y.~Slimani \emph{et~al.} 2019.
\newblock Tracking community evolution in social networks: A survey.
\newblock \emph{Information Processing \& Management}
  \textbf{56}(3):1084--1102.

\bibitem[{Dinh \emph{et~al.}(2009)Dinh, Xuan and Thai}]{dinh2009towards}
Dinh, T.~N., Y.~Xuan and M.~T. Thai. 2009.
\newblock Towards social-aware routing in dynamic communication networks.
\newblock In: \emph{2009 IEEE 28th International Performance Computing and
  Communications Conference}, pages 161--168. IEEE.

\bibitem[{Farina(2015)}]{farina2015congressional}
Farina, C.~R. 2015.
\newblock Congressional polarization: terminal constitutional dysfuction.
\newblock \emph{Colum L Rev} \textbf{115}:1689.

\bibitem[{Ferrari \emph{et~al.}(2019)Ferrari, Lindholm and
  K{\"o}nig}]{ferrari2019fitness}
Ferrari, M., A.~K. Lindholm and B.~K{\"o}nig. 2019.
\newblock Fitness consequences of female alternative reproductive tactics in
  house mice (Mus musculus domesticus).
\newblock \emph{The American Naturalist} \textbf{193}(1):106--124.

\bibitem[{Fortunato(2010)}]{fortunato2010community}
Fortunato, S. 2010.
\newblock Community detection in graphs.
\newblock \emph{Physics reports} \textbf{486}(3-5):75--174.

\bibitem[{Fortunato and Hric(2016)}]{fortunato2016community}
Fortunato, S. and D.~Hric. 2016.
\newblock Community detection in networks: A user guide.
\newblock \emph{Physics Reports} \textbf{659}:1--44.

\bibitem[{Geiger \emph{et~al.}(2018)Geiger, S{\'a}nchez-Villagra and
  Lindholm}]{geiger2018longitudinal}
Geiger, M., M.~R. S{\'a}nchez-Villagra and A.~K. Lindholm. 2018.
\newblock A longitudinal study of phenotypic changes in early domestication of
  house mice.
\newblock \emph{Royal Society open science} \textbf{5}(3):172099.

\bibitem[{Holme(2015)}]{holme2015modern}
Holme, P. 2015.
\newblock Modern temporal network theory: a colloquium.
\newblock \emph{The European Physical Journal B} \textbf{88}(9):234.

\bibitem[{Jaccard(1901)}]{jaccard1901distribution}
Jaccard, P. 1901.
\newblock Distribution de la flore alpine dans le bassin des Dranses et dans
  quelques r{\'e}gions voisines.
\newblock \emph{Bull Soc Vaudoise Sci Nat} \textbf{37}:241--272.

\bibitem[{Jain \emph{et~al.}(1999)Jain, Murty and Flynn}]{jain1999data}
Jain, A.~K., M.~N. Murty and P.~J. Flynn. 1999.
\newblock Data clustering: a review.
\newblock \emph{ACM computing surveys (CSUR)} \textbf{31}(3):264--323.

\bibitem[{Kaufman and Rousseeuw(2009)}]{kaufman2009finding}
Kaufman, L. and P.~J. Rousseeuw. 2009.
\newblock \emph{Finding groups in data: an introduction to cluster analysis},
  volume 344.
\newblock John Wiley \& Sons.

\bibitem[{K{\"o}nig \emph{et~al.}(2012)K{\"o}nig, Lindholm
  \emph{et~al.}}]{konig2012complex}
K{\"o}nig, B., A.~Lindholm \emph{et~al.} 2012.
\newblock The complex social environment of female house mice (Mus domesticus).
\newblock \emph{Evolution of the house mouse} \textbf{114}:134.

\bibitem[{K{\"o}nig \emph{et~al.}(2015)K{\"o}nig, Lindholm, Lopes
  \emph{et~al.}}]{konig2015system}
K{\"o}nig, B., A.~K. Lindholm, P.~C. Lopes \emph{et~al.} 2015.
\newblock A system for automatic recording of social behavior in a free-living
  wild house mouse population.
\newblock \emph{Animal Biotelemetry} \textbf{3}(1):39.

\bibitem[{Lewis \emph{et~al.}(2018)Lewis, Poole, Rosenthal
  \emph{et~al.}}]{lewis2018voteview}
Lewis, J.~B., K.~Poole, H.~Rosenthal \emph{et~al.} 2018.
\newblock Voteview: Congressional roll-call votes database.
\newblock \emph{See https://voteview com/(accessed 27 July 2018)} .

\bibitem[{Liechti \emph{et~al.}(submitted)Liechti, Qian, K\"{o}nig, Bonhoeffer}]{liechti2020ff}
  Liechti, J.~I., B.~Qian, B.~K\"{o}nig and S.~Bonhoeffer \emph{submitted}.
\newblock Contact patterns reveal a stable dynamic community structure with
  fission-fusion dynamics in wild house mice.

\bibitem[{Mucha \emph{et~al.}(2010)Mucha, Richardson, Macon
  \emph{et~al.}}]{mucha2010community}
Mucha, P.~J., T.~Richardson, K.~Macon \emph{et~al.} 2010.
\newblock Community structure in time-dependent, multiscale, and multiplex
  networks.
\newblock \emph{Science} \textbf{328}(5980):876--878.

\bibitem[{Palla \emph{et~al.}(2007)Palla, Barab{\'a}si and
  Vicsek}]{palla2007quantifying}
Palla, G., A.-L. Barab{\'a}si and T.~Vicsek. 2007.
\newblock Quantifying social group evolution.
\newblock \emph{Nature} \textbf{446}(7136):664.

\bibitem[{Palla \emph{et~al.}(2005)Palla, Der{\'e}nyi, Farkas
  \emph{et~al.}}]{palla2005uncovering}
Palla, G., I.~Der{\'e}nyi, I.~Farkas \emph{et~al.} 2005.
\newblock Uncovering the overlapping community structure of complex networks in
  nature and society.
\newblock \emph{Nature} \textbf{435}(7043):814.

\bibitem[{Rosenberg and Hirschberg(2007)}]{rosenberg2007v}
Rosenberg, A. and J.~Hirschberg. 2007.
\newblock V-measure: A conditional entropy-based external cluster evaluation
  measure.
\newblock In: \emph{Proceedings of the 2007 joint conference on empirical
  methods in natural language processing and computational natural language
  learning (EMNLP-CoNLL)}.

\bibitem[{Rosvall and Bergstrom(2008)}]{rosvall2008maps}
Rosvall, M. and C.~T. Bergstrom. 2008.
\newblock Maps of random walks on complex networks reveal community structure.
\newblock \emph{Proceedings of the National Academy of Sciences}
  \textbf{105}(4):1118--1123.

\bibitem[{Rosvall and Bergstrom(2010)}]{rosvall2010mapping}
Rosvall, M. and C.~T. Bergstrom. 2010.
\newblock Mapping change in large networks.
\newblock \emph{PloS one} \textbf{5}(1):e8694.

\bibitem[{Sun \emph{et~al.}(2007)Sun, Faloutsos, Faloutsos
  \emph{et~al.}}]{sun2007graphscope}
Sun, J., C.~Faloutsos, C.~Faloutsos \emph{et~al.} 2007.
\newblock Graphscope: parameter-free mining of large time-evolving graphs.
\newblock In: \emph{Proceedings of the 13th ACM SIGKDD international conference
  on Knowledge discovery and data mining}, pages 687--696. ACM.

\bibitem[{Waugh \emph{et~al.}(2009)Waugh, Pei, Fowler
  \emph{et~al.}}]{waugh2009party}
Waugh, A.~S., L.~Pei, J.~H. Fowler \emph{et~al.} 2009.
\newblock Party polarization in congress: A network science approach.

\end{thebibliography}

{\small
  \begin{thebibliography}{3}
\providecommand{\natexlab}[1]{#1}

\bibitem[{Hunter(2007)}]{hunter2007matplotlib}
Hunter, J.~D. 2007.
\newblock Matplotlib: A 2D graphics environment.
\newblock \emph{Computing in science \& engineering} \textbf{9}(3):90.

\bibitem[{Jaccard(1901)}]{jaccard1901distribution}
Jaccard, P. 1901.
\newblock Distribution de la flore alpine dans le bassin des Dranses et dans
  quelques r{\'e}gions voisines.
\newblock \emph{Bull Soc Vaudoise Sci Nat} \textbf{37}:241--272.

\bibitem[{Liechti(2020)}]{jonas_i_liechti_2020_3678351}
Liechti, J.~I. 2020.
\newblock github.com/j-i-l/pyAlluv - v0.2 - Matplotlib based alluvial diagrams
  in python.

\end{thebibliography}

  }
\end{document}